\documentclass[10pt,journal,compsoc]{IEEEtran}

\usepackage{amsmath}
\usepackage{diagbox}
\usepackage{graphics}
\usepackage{subfigure}
\usepackage{epsfig}
\usepackage{threeparttable}
\usepackage{xspace}

\usepackage{siunitx}

\usepackage{graphicx}
\usepackage{amssymb}
\usepackage{threeparttable}
\usepackage{color}
\usepackage[normalem]{ulem}
\usepackage{multirow}
\usepackage{float}
\usepackage{amsfonts}
\usepackage{bm}
\usepackage{array}
\usepackage{colortbl}
\usepackage{diagbox}
\usepackage{rotating}
\usepackage{booktabs}
\usepackage{overpic}
\usepackage{enumitem}
\usepackage{colortbl}
\usepackage{dblfloatfix}

\usepackage[ruled, vlined]{algorithm2e}
\usepackage[textsize=tiny]{todonotes}
\PassOptionsToPackage{table,x11names}{xcolor}  
\usepackage{xcolor}  
\usepackage{fontawesome}
\usepackage{mdframed}
\usepackage{pifont}
\usepackage{tabularx} 
\usepackage[export]{adjustbox}
\usepackage{threeparttable}
\usepackage{subcaption}
\usepackage{multirow}
\usepackage{enumitem}
\usepackage{newfloat}
\usepackage{listings}
\usepackage{colortbl}
\usepackage{arydshln}
\usepackage{amsfonts}
\usepackage{makecell}
\usepackage{caption}
\usepackage{cancel}
\usepackage{wrapfig}
\usepackage{amssymb}
\usepackage{ragged2e}
\usepackage{bbding}
\usepackage{float}
\usepackage{bbm}
\usepackage{bm}
\usepackage{mathrsfs} 

\newcommand{\myhyperlink}[3][black]{\hyperlink{#2}{\color{#1}{#3}}}
\usepackage{colortbl}
\definecolor{lightgray}{gray}{.9}
\definecolor{deepgray}{gray}{.8}

\usepackage{threeparttable}
\newcolumntype{I}{!{\vrule width 1pt}}
\makeatletter
\newcommand{\thickhline}{%
    \noalign {\ifnum 0=`}\fi \hrule height 1pt
    \futurelet \reserved@a \@xhline
}
\makeatother
\definecolor{mygray}{gray}{.9}

\usepackage[pagebackref=false,breaklinks=true,colorlinks,bookmarks=false]{hyperref}

\definecolor{mydarkdarkred}{RGB}{182, 58, 43} 
\definecolor{mydarkdarkred2}{RGB}{180, 92, 67} 
\definecolor{mydarkred}{RGB}{245, 220, 215}  
\definecolor{mylightred}{RGB}{251, 244, 242}  

\definecolor{mydarkdarkblue}{RGB}{17, 24, 129} 
\definecolor{mydarkdarkblue2}{RGB}{92, 108, 165} 
\definecolor{mydarkblue}{RGB}{211, 218, 234} 
\definecolor{mylightblue}{RGB}{241, 244, 250}

\definecolor{mydarkdarkgreen}{RGB}{93, 150, 74} 
\definecolor{mydarkgreen}{RGB}{216, 233, 199}  
\definecolor{mylightgreen}{RGB}{245, 249, 241}

\definecolor{darkred}{RGB}{139,0,0}  

\definecolor{mydarkyellow}{RGB}{216, 214, 196}   
\definecolor{mylightyellow}{RGB}{245, 245, 240} 

\definecolor{mygray}{gray}{.9}
\definecolor{mygreen}{RGB}{93,173,85}
\definecolor{myorange}{RGB}{233,144,61}
\definecolor{azure}{rgb}{0.0, 0.5, 1.0}
\definecolor{gray}{rgb}{0.3, 0.3, 0.3}
\definecolor{DarkGreen}{RGB}{42,110,63}
\definecolor{DarkBlue}{RGB}{64,101,149}

\definecolor{mylightgray}{RGB}{249, 249, 249}

\newcommand{\pub}[1]{{\color{gray}{\footnotesize{[{#1}]}}}}

\newcommand{\tworowup}[2]{
\begin{tabular}{@{}c@{}}
{#1}  \\
\fontsize{6.5pt}{1em}\selectfont\color{myorange}{$\uparrow$ \textbf{#2}}
\end{tabular}
}

\newcommand{\tworowdown}[2]{
\begin{tabular}{@{}c@{}}
{#1}  \\
\fontsize{6.5pt}{1em}\selectfont\color{mygreen}{$\downarrow$ \textbf{#2}}
\end{tabular}
}

\newtheorem{theorem}{Theorem}[section]

\newtheorem{definition}[theorem]{Definition}

\usepackage[capitalize]{cleveref}
\crefname{section}{Sec.}{Secs.}
\crefname{table}{Tab.}{Tabs.}
\crefname{section}{§}{§§}

\makeatletter
\DeclareRobustCommand\onedot{\futurelet\@let@token\@onedot}
\def\@onedot{\ifx\@let@token.\else.\null\fi\xspace}

\def\eg{\emph{e.g}\onedot} 
\def\ie{\emph{i.e}\onedot}

\makeatletter
\newenvironment{fullitemize}
{
\begin{itemize}[leftmargin=*]
\setlength{\itemsep}{3pt}
\setlength{\parsep}{-5pt}
\setlength{\parskip}{-3pt}
\setlength{\leftmargin}{-10pt}
}
{
\end{itemize}
}
\makeatother

\begin{document}
\title{Keeping Yourself is Important in Downstream Tuning Multimodal Large Language Model}
\author{
Wenke Huang$^*$,  Jian Liang$^*$, Xianda Guo$^*$, Yiyang Fang$^*$, Guancheng Wan$^*$, Xuankun Rong\\ 
Chi Wen, Zekun Shi, Qingyun Li, Didi Zhu, Yanbiao Ma, Ke Liang, Bin Yang, He Li,  \\ 
Jiawei Shao, Mang Ye$^\dagger$, Bo Du$^\dagger$
\IEEEcompsocitemizethanks{
\IEEEcompsocthanksitem 
Wenke Huang, Jian Liang, Xianda Guo, Yiyang Fang, Xuankun Rong, Zekun Shi, Bin Yang, He Li, Mang Ye, and Bo Du are with the School of Computer Science, Wuhan University, Wuhan, 430072, China. \protect
E-mail:\{wenkehuang, yemang, dubo\}@whu.edu.cn
\IEEEcompsocthanksitem 
Qingyun Li is with the school of Electronics and Information Engineering, Harbin Institute of Technology, Harbin, 150001, China. 
\IEEEcompsocthanksitem 
Didi Zhu is with the Department of Computer Science and Technology, Zhejiang University, Hangzhou, 310058, China
\IEEEcompsocthanksitem
Yanbiao Ma is with the School of Artificial Intelligence, Xidian University, Xian, 710071, China
\IEEEcompsocthanksitem 
Ke Liang is with the College of Computer Science and Technology, National University of Defense Technology, Changsha, 410073, China. 
\IEEEcompsocthanksitem
Jiawei Shao is with the Institute of Artificial Intelligence (TeleAI), China
\IEEEcompsocthanksitem $^*$ Indicates equal contribution.
\IEEEcompsocthanksitem Corresponding Authors: Mang Ye and Bo Du
}
}
\markboth{Keeping Yourself is Important in Downstream
Tuning Multimodal Large Language Model}%
{Shell \MakeLowercase{\textit{et al.}}: Bare Demo of IEEEtran.cls for Journals}

\IEEEtitleabstractindextext{
\begin{abstract}
Multi-modal Large Language Models (MLLMs) integrate visual and linguistic reasoning to address complex tasks such as image captioning and visual question answering. While MLLMs demonstrate remarkable versatility, MLLMs appears limited performance on special applications. But tuning MLLMs for downstream tasks encounters two key challenges: Task-Expert Specialization, where distribution shifts between pre-training and target datasets constrain target performance, and Open-World Stabilization, where catastrophic forgetting erases the model general knowledge. In this work, we systematically review recent advancements in MLLM tuning methodologies, classifying them into three paradigms: (I) Selective Tuning, (II) Additive Tuning, and (III) Reparameterization Tuning. Furthermore, we benchmark these tuning strategies across popular MLLM architectures and diverse downstream tasks to establish standardized evaluation analysis and systematic tuning principles. Finally, we highlight several open challenges in this domain and propose future research directions. To facilitate ongoing progress in this rapidly evolving field, we provide a public repository that continuously tracks developments: \url{https://github.com/WenkeHuang/Awesome-MLLM-Tuning}.
\end{abstract}
}

\maketitle
\IEEEdisplaynontitleabstractindextext
\IEEEpeerreviewmaketitle
\newcommand{\digits}{{Digits}}
\newcommand{\mnist}{{MNIST}}
\newcommand{\avmnist}{{AVMNIST}}
\newcommand{\mnistabbrv}{{M}}
\newcommand{\usps}{{USPS}}
\newcommand{\uspsabbrv}{{U}}
\newcommand{\svhn}{{SVHN}}
\newcommand{\svhnabbrv}{{Sv}}
\newcommand{\syn}{{SYN}}
\newcommand{\synabbrv}{{Sy}}

\newcommand{\officecaltech}{{Office Caltech}}
\newcommand{\caltech}{{Caltech}}
\newcommand{\caltechabbrv}{{Ca}}
\newcommand{\amazon}{{Amazon}}
\newcommand{\amazonabbrv}{{Am}}
\newcommand{\webcam}{{Webcam}}
\newcommand{\domainnet}{{DomainNet}}
\newcommand{\webcamabbrv}{{W}}
\newcommand{\dslr}{{DSLR}}
\newcommand{\dslrabbrv}{{D}}

\newcommand{\refcoco}{{RefCOCO}}
\newcommand{\icfg}{{ICFG}}
\newcommand{\caltechtfs}{{Caltech-256}}
\newcommand{\officeto}{{Office31}}

\newcommand{\officehome}{{Office-Home}}
\newcommand{\art}{{Art}}
\newcommand{\artabbrv}{{Ar}}
\newcommand{\clipart}{{Clipart}}
\newcommand{\clipartabbrv}{{Cl}}
\newcommand{\product}{{Product}}
\newcommand{\productabbrv}{{Pr}}
\newcommand{\realworld}{{Real World}}
\newcommand{\realworldabbrv}{{RW}}

\newcommand{\pacs}{{PACS}}
\newcommand{\photo}{{Photo}}
\newcommand{\photoabbrv}{{P}}
\newcommand{\artpainting}{{Art Painting}}
\newcommand{\artpaintingabbrv}{{AP}}
\newcommand{\cartoon}{{Cartoon}}
\newcommand{\cartoonabbrv}{{Ct}}
\newcommand{\sketch}{{Sketch}}
\newcommand{\sketchabbrv}{{Sk}}
\newcommand{\infographdomain}{{Infograph}}
\newcommand{\infographdomainabbrv}{{I}}
\newcommand{\painting}{{Painting}}
\newcommand{\paintingabbrv}{{Pa}}
\newcommand{\Quickdraw}{{Quickdraw}}
\newcommand{\Quickdrawabbrv}{{Q}}
\newcommand{\Real}{{Real}}
\newcommand{\Realabbrv}{{R}}

\newcommand{\gqa}{{GQA}}
\newcommand{\textvqa}{{TextVQA}}
\newcommand{\ocrvqa}{{OCRVQA}}
\newcommand{\okvqa}{{OKVQA}}
\newcommand{\vizwiz}{{Vizwiz}}
\newcommand{\scienceqa}{{ScienceQA}}
\newcommand{\flickrthirtyk}{{Flickr30k}}
\newcommand{\iconqa}{{IconQA}}
\newcommand{\vqavtwo}{{VQAV2}}
\newcommand{\cococap}{{COCO-Cap}}
\newcommand{\mme}{{MME}}
\newcommand{\pope}{{POPE}}

\newcommand{\vlcs}{{VLCS}}
\newcommand{\labelme}{{Labelme}}
\newcommand{\labelmeabbrv}{{L}}
\newcommand{\pascalabbrv}{{Pa}}
\newcommand{\sun}{{Sun}}
\newcommand{\sunabbrv}{{Su}}

\newcommand{\cifar}{{Cifar}}
\newcommand{\weperson}{{WePerson}}
\newcommand{\imagenet}{{ImageNet}}
\newcommand{\cifarhun}{{Cifar-100}}
\newcommand{\cifarten}{{Cifar-10}}
\newcommand{\tyimg}{{Tiny-ImageNet}}
\newcommand{\coco}{{COCO}}
\newcommand{\market}{{Market1501}}
\newcommand{\fashionmnist}{{Fashion-MNIST}}
\newcommand{\cremad}{{CREAM-D}}
\newcommand{\ave}{{AVE}}
\newcommand{\sswsixty}{{SSW60}}
\newcommand{\cmumosei}{{CMU-MOSEI}}

\newcommand{\lasso}{{Lasso}}
\newcommand{\pls}{{PLS}}
\newcommand{\olsregression}{{OLS}}
\newcommand{\mlp}{{MLP}}
\newcommand{\gru}{{GRU}}
\newcommand{\lstm}{{LSTM}}
\newcommand{\ridge}{{Ridge}}
\newcommand{\linereg}{{LineReg}}
\newcommand{\elasticnet}{{Elastic Net}}
\newcommand{\rsvqa}{{RSVQA}}

\newcommand{\simplecnn}{{SimpleCNN}}
\newcommand{\resnet}{{ResNet}}
\newcommand{\resnetten}{{ResNet-10}}
\newcommand{\resnettwelve}{{ResNet-12}}
\newcommand{\resneteighteen}{{ResNet-18}}
\newcommand{\resnettwenty}{{ResNet-20}}
\newcommand{\resnetthirtyfour}{{ResNet-34}}
\newcommand{\resnetfifty}{{ResNet-50}}
\newcommand{\resnetonezeroone}{{ResNet-101}}
\newcommand{\resnetonefivetwo}{{ResNet-152}}

\newcommand{\resnext}{{ResNeXt}}
\newcommand{\ResNextFourEleven}{{ResNeXt4-11}}
\newcommand{\mobilenet}{{MobileNet}}
\newcommand{\efficientnet}{{EfficientNet}}
\newcommand{\densenet}{{DenseNet}}
\newcommand{\convnet}{{Convnet}}
\newcommand{\googlenet}{{GoogLeNet}}
\newcommand{\fbnet}{{Fbnet}}

\newcommand{\sgd}{{SGD}}
\newcommand{\adam}{{Adam}}

\newcommand{\finch}{{Finch}}
\newcommand{\hac}{{HAC}}
\newcommand{\kmeans}{{Kmeans}}
\newcommand{\cka}{{CKA}}
\newcommand{\tsne}{{t-SNE}}
\newcommand{\dbscan}{{DBSCAN}}
\newcommand{\hnne}{{h-NNE}}

\newcommand{\ce}{{CE Loss}}
\newcommand{\kl}{{KL Loss}}
\newcommand{\triplet}{{Triplet Loss}}
\newcommand{\normface}{{Normface}}
\newcommand{\sphereface}{{Sphereface}}
\newcommand{\rce}{{RCE}}
\newcommand{\proxyanchor}{{Proxy-Anchor}}
\newcommand{\atk}{{$AT_k$}}
\newcommand{\centerloss}{{Center}}
\newcommand{\largemargin}{{L-Softmax}}
\newcommand{\arcface}{{ArcFace}}
\newcommand{\polyloss}{{PolyLoss}}

\newcommand{\attention}{{Attention}}
\newcommand{\bert}{{BERT}}
\newcommand{\vit}{{ViT}}
\newcommand{\aft}{{AFT}}
\newcommand{\cpvt}{{CPVT}}
\newcommand{\alphaedit}{{AlphaEdit}}
\newcommand{\deit}{{DeiT}}
\newcommand{\convit}{{ConViT}}
\newcommand{\ceit}{{CeiT}}
\newcommand{\cogvlm}{{CogVLM}}
\newcommand{\localvit}{{LocalViT}}
\newcommand{\swin}{{Swin}}
\newcommand{\child}{{CHILD}}
\newcommand{\sat}{{SAT}}
\newcommand{\atmmerge}{{ATM}}
\newcommand{\cvt}{{CvT}}
\newcommand{\vhm}{{VHM}}
\newcommand{\spider}{{SPIDER}}
\newcommand{\cofitune}{{CoFiTune}}
\newcommand{\galore}{{GaLore}}
\newcommand{\id}{{ID}}
\newcommand{\lors}{{LoRS}}
\newcommand{\vacode}{{VaCoDe}}
\newcommand{\arf}{{ARF}}
\newcommand{\lorarar}{{LoRA.rar}}
\newcommand{\janus}{{Janus}}
\newcommand{\cart}{{CART}}
\newcommand{\fastvlm}{{FastVLM}}
\newcommand{\minigptfour}{{MiniGPT4}}
\newcommand{\fishdip}{{FISHDIP}}
\newcommand{\dlc}{{DLC}}
\newcommand{\cambrianone}{{Cambrian-1}}
\newcommand{\framefusion}{{FrameFusion}}
\newcommand{\autoprompt}{{AutoPrompt}}
\newcommand{\lpaqa}{{LPAQA}}
\newcommand{\rose}{{ROSE}}
\newcommand{\icd}{{ICD}}
\newcommand{\unilm}{{UNILM}}
\newcommand{\vcd}{{VCD}}
\newcommand{\llavagrounding}{{LLaVA-Grounding}}
\newcommand{\visionzip}{{VisionZip}}
\newcommand{\pmodmllm}{{p-MoD}}
\newcommand{\cswin}{{CSWin}}
\newcommand{\lavin}{{LaVIN}}
\newcommand{\pvp}{{PVP}}
\newcommand{\lorascruplt}{{LoRASculpt}}
\newcommand{\victor}{{Victor}}
\newcommand{\llavauhd}{{LLaVA-UHD}}
\newcommand{\corda}{{CorDA}}
\newcommand{\woodpecker}{{Woodpecker}}
\newcommand{\bioclip}{{BioCLIP}}
\newcommand{\marine}{{MARINE}}
\newcommand{\pali}{{PaLI}}
\newcommand{\sfllava}{{SF-LLaVA}}
\newcommand{\ritual}{{RITUAL}}
\newcommand{\setar}{{SeTAR}}
\newcommand{\ctts}{{CT$^2$S}}
\newcommand{\dor}{{DOR}}
\newcommand{\avisc}{{AvisC}}
\newcommand{\alcd}{{ALCD}}
\newcommand{\mustdrop}{{MustDrop}}
\newcommand{\pai}{{PAI}}
\newcommand{\mtlora}{{MTLoRA}}
\newcommand{\bilora}{{BiLoRA}}
\newcommand{\laplacelora}{{Lap-LoRA}}
\newcommand{\hio}{{HIO}}
\newcommand{\gtp}{{GTP}}
\newcommand{\moslora}{{MoSLoRA}}
\newcommand{\patch}{{PATCH}}
\newcommand{\vdgd}{{VDGD}}
\newcommand{\miniplm}{{MiniPLM}}
\newcommand{\llavaalign}{{LLaVA-Align}}
\newcommand{\hydralora}{{HydraLoRA}}
\newcommand{\tcprompt}{{TCP}}
\newcommand{\melora}{{MeLoRA}}
\newcommand{\reslora}{{ResLoRA}}
\newcommand{\vera}{{VERA}}
\newcommand{\atkd}{{ATKD}}
\newcommand{\minillm}{{MiniLLM}}
\newcommand{\fdivergence}{\textit{f}-DISTILL}
\newcommand{\emrmerging}{{EMRMerge}}
\newcommand{\smilellm}{{SMILE}}
\newcommand{\teamlora}{{TeamLoRA}}
\newcommand{\hiddenkey}{{HiddenKey}}
\newcommand{\taskarithmetic}{{Task-Arithmetic}}
\newcommand{\sparsegpt}{{SparseGPT}}
\newcommand{\twinmerge}{{Twin}}
\newcommand{\tac}{{TAC}}
\newcommand{\besa}{{BESA}}
\newcommand{\dam}{{DAM}}
\newcommand{\dcc}{{DCC}}
\newcommand{\ipt}{{IPT}}
\newcommand{\sparsevlm}{{SpareseVLM}}
\newcommand{\ties}{{TIES}}
\newcommand{\vila}{{VILA}}
\newcommand{\lorahub}{{LoraHub}}
\newcommand{\llavaphi}{{LLaVA-$\phi$}}
\newcommand{\vilatwo}{{VILA$^2$}}
\newcommand{\llamavid}{{LLaMA-VID}}
\newcommand{\bliva}{{BLIVA}}
\newcommand{\sdft}{{SDFT}}
\newcommand{\argue}{{ArGue}}
\newcommand{\llava}{{LLaVA}}
\newcommand{\llavaov}{{LLaVA-OV}}
\newcommand{\dora}{{DoRA}}
\newcommand{\mofo}{{MoFO}}
\newcommand{\lisa}{{LISA}}
\newcommand{\llavahr}{{LLaVA-HR}}
\newcommand{\smop}{{SMoP}}
\newcommand{\loraplus}{{LoRA+}}
\newcommand{\sphinx}{{SPHINX}}
\newcommand{\honeybee}{Honeybee}
\newcommand{\remedy}{{REMEDY}}
\newcommand{\vtw}{{VTW}}
\newcommand{\mpvit}{{MPViT}}
\newcommand{\cropa}{{CroPA}}
\newcommand{\spu}{{SPU}}
\newcommand{\loramoe}{{LoRAMoE}}
\newcommand{\lmbff}{{LM-BFF}}
\newcommand{\llama}{{LLaMA}}
\newcommand{\vicuna}{{Vicuna}}
\newcommand{\cpr}{{CPR}}
\newcommand{\spt}{{SPT}}
\newcommand{\flora}{{Flora}}
\newcommand{\coprompt}{{CoPrompt}}
\newcommand{\dare}{{DARE}}
\newcommand{\lth}{{LTH}}
\newcommand{\milora}{{MiLoRA}}
\newcommand{\tasl}{{TaSL}}
\newcommand{\fish}{{FISH}}
\newcommand{\wanda}{{Wanda}}
\newcommand{\maskllm}{{MaskLLM}}
\newcommand{\mthreeid}{{M3ID}}
\newcommand{\rlcf}{{RLCF}}
\newcommand{\clora}{{CLoRA}} 
\newcommand{\stprompt}{{STP}}
\newcommand{\neglabel}{{NegLabel}}
\newcommand{\restore}{{RESTORE}}
\newcommand{\prolora}{{PRoLoRA}}
\newcommand{\mixlora}{{MixLoRA}}
\newcommand{\dpl}{{DPL}}
\newcommand{\ptp}{{PTP}}
\newcommand{\swarm}{{SWARM}}
\newcommand{\damp}{{DAMP}}
\newcommand{\dpr}{{DPR}}
\newcommand{\magpurne}{{MagPurne}}
\newcommand{\saco}{{SaCo}}
\newcommand{\cpt}{{CPT}}
\newcommand{\cmpa}{{CMPA}}
\newcommand{\vars}{{VARS}}
\newcommand{\krona}{{Krona}}
\newcommand{\coda}{{CoDA}}
\newcommand{\clipvip}{{CLIP-ViP}}
\newcommand{\vipllava}{{ViP-LLaVA}}
\newcommand{\ciat}{{CIAT}}
\newcommand{\sift}{{SIFT}}
\newcommand{\dits}{{DiTs}}
\newcommand{\aprompt}{{Aprompt}}
\newcommand{\absvit}{{AbSViT}}
\newcommand{\vlp}{{Vision-Language Pre-trained Models}}
\newcommand{\vlpab}{{VLPM}}
\newcommand{\san}{{SAN}}
\newcommand{\supermerge}{{SUPERMERGE}}
\newcommand{\clapaudio}{{CLAP}}
\newcommand{\pissa}{{PiSSA}}
\newcommand{\rlt}{{RLT}}
\newcommand{\cascadeclip}{{Cascade-CLIP}}
\newcommand{\declip}{{DeCLIP}}
\newcommand{\imagebind}{{ImageBind}}
\newcommand{\uavod}{{UAV-OD}}
\newcommand{\clip}{{CLIP}}
\newcommand{\bop}{{BoP}}
\newcommand{\bicro}{{BiCro}}
\newcommand{\defo}{{DeFO}}
\newcommand{\llavamerge}{{LLaVA-PruMerge}}
\newcommand{\flamingo}{{Flamingo}}
\newcommand{\ssam}{{SSAM}}
\newcommand{\alip}{{ALIP}}
\newcommand{\gps}{{GPS}}
\newcommand{\prefixtuning}{{PrefixTune}}
\newcommand{\npt}{{NPT}}
\newcommand{\dac}{{DAC}}
\newcommand{\fsam}{{FSAM}}
\newcommand{\livt}{{LiVT}}
\newcommand{\scmoe}{{SCMoE}}
\newcommand{\instructblip}{{InstructBLIP}}
\newcommand{\blip}{{BLIP}}
\newcommand{\bliptwo}{{BLIP-2}}
\newcommand{\qwenvl}{{QWen-VL}}
\newcommand{\stwo}{{S$^{2}$}}
\newcommand{\alignVL}{{ALIGN}}
\newcommand{\ramt}{{R-AMT}}
\newcommand{\adamvmoe}{{AdaMVMoE}}
\newcommand{\tpt}{{TPT}}
\newcommand{\graphadapter}{{GraphAdapter}}
\newcommand{\lingualsmoe}{{Lingual-SMoE}}
\newcommand{\tai}{{TaI}}
\newcommand{\dapt}{{DAPT}}
\newcommand{\dualprompt}{{DualPrompt}}
\newcommand{\dept}{{DePT}}
\newcommand{\losparse}{{LoSparse}}
\newcommand{\fuller}{{FULLER}}
\newcommand{\depthclip}{{DepthCLIP}}
\newcommand{\vlab}{{VLAB}}
\newcommand{\lpt}{{LPT}}
\newcommand{\tvp}{{TVP}}
\newcommand{\promptkd}{{PromptKD}}
\newcommand{\metatrans}{{Meta-Transformer}}
\newcommand{\gasam}{{GA-SAM}}
\newcommand{\sprompt}{{SP}}
\newcommand{\promptstyler}{{PromptStyler}}
\newcommand{\umt}{{UMT}}
\newcommand{\pego}{{PEGO}}
\newcommand{\plot}{{PLOT}}
\newcommand{\doprompt}{{DoPrompt}}
\newcommand{\misa}{{MISA}}
\newcommand{\protext}{{ProText}}
\newcommand{\distllm}{{DISTLLM}}
\newcommand{\vipt}{{ViPT}}
\newcommand{\kgcoop}{{KgCoOp}}
\newcommand{\bridge}{{Birdge}}
\newcommand{\dferc}{{DFERC}}
\newcommand{\xprompt}{{Xprompt}}
\newcommand{\tailor}{{Tailor}}
\newcommand{\dualcoopplus}{{DualCoOP++}}
\newcommand{\upt}{{UPT}}
\newcommand{\sharelora}{{ShareLoRA}}
\newcommand{\itwomcl}{{I$^2$MCL}}
\newcommand{\gblending}{{G-Blending}}
\newcommand{\greedy}{{Greedy}}
\newcommand{\mmanet}{{MMANet}}
\newcommand{\grda}{{GRDA}}
\newcommand{\lmf}{{LMF}}
\newcommand{\kapt}{{KAPT}}
\newcommand{\mfm}{{MFM}}
\newcommand{\etwovpt}{{E$^2$VPT}}
\newcommand{\audioclip}{{AudioCLIP}}
\newcommand{\hotprotein}{{HotProtein}}
\newcommand{\prac}{{PRAC}}
\newcommand{\promptsrc}{{PromptSRC}}
\newcommand{\diffpurning}{{DiffPurning}}
\newcommand{\prograd}{{ProGrad}}
\newcommand{\pcb}{{PCB}}
\newcommand{\bike}{{BIKE}}
\newcommand{\maple}{{MaPLe}}
\newcommand{\fdmer}{{FDMER}}
\newcommand{\mtwopt}{{M2PT}}
\newcommand{\film}{{FiLM}}
\newcommand{\laclip}{{LaCLIP}}
\newcommand{\selfmm}{{Self-MM}}
\newcommand{\conc}{{Concatenation}}
\newcommand{\summ}{{Summation}}
\newcommand{\clipreid}{{CLIP-REID}}
\newcommand{\gated}{{Gated}}
\newcommand{\capforvideo}{{Cap4Video}}
\newcommand{\agm}{{AGM}}
\newcommand{\icode}{{i-Code}}
\newcommand{\msrl}{{MSRL}}
\newcommand{\mslr}{{MSLR}}
\newcommand{\fagm}{{FAGM}}
\newcommand{\taslplus}{{TaSLPLUS}}
\newcommand{\dualcoop}{{DualCoOp}}
\newcommand{\saf}{{SAF}}
\newcommand{\shape}{{SHAPE}}
\newcommand{\rigorllm}{{RigorLLM}}
\newcommand{\aptm}{{APTM}}
\newcommand{\fishr}{{Fishr}}
\newcommand{\soho}{{SOHO}}
\newcommand{\ijepa}{{I-JEPA}}
\newcommand{\pmr}{{PMR}}
\newcommand{\smil}{{SMIL}}
\newcommand{\man}{{MAN}}
\newcommand{\mbt}{{MBT}}
\newcommand{\dmd}{{DMD}}
\newcommand{\pmf}{{PMF}}
\newcommand{\tipadater}{{Tip-Adapter}}
\newcommand{\clipadapter}{{CLIP-Adapter}}
\newcommand{\coop}{{CoOP}}
\newcommand{\softmask}{{SoftMask}}
\newcommand{\clippo}{{CLIPPO}}
\newcommand{\cocoop}{{CoCoOp}}
\newcommand{\cafo}{{CaFo}}
\newcommand{\pcme}{{PCME}}
\newcommand{\ape}{{APE}}
\newcommand{\ogmge}{{OGMGE}}
\newcommand{\vificlip}{{ViFi-CLIP}}
\newcommand{\fdt}{{FDT}}
\newcommand{\hpt}{{HPT}}
\newcommand{\shaspec}{{ShaSpec}}
\newcommand{\cleanclip}{{CleanCLIP}}
\newcommand{\fourierft}{{FourierFT}}
\newcommand{\vilt}{{ViLT}}
\newcommand{\metaadapter}{{MetaAdapter}}
\newcommand{\gmc}{{GMC}}
\newcommand{\denseclip}{{DenseCLIP}}
\newcommand{\chils}{{CHiLS}}
\newcommand{\mtwofnet}{{M2FNet}}
\newcommand{\mmcosine}{{MMCosine}}
\newcommand{\lora}{{LoRA}}
\newcommand{\offsitetuning}{{Offsite-Tuning}}
\newcommand{\adapter}{{Adapters}}
\newcommand{\pst}{{PST}}
\newcommand{\lst}{{LST}}
\newcommand{\ltsft}{{LT-SFT}}
\newcommand{\moefication}{{MoEfication}}
\newcommand{\toast}{{TOAST}}
\newcommand{\fastv}{{FastV}}
\newcommand{\adalora}{{AdaLoRA}}
\newcommand{\pafi}{{PaFi}}
\newcommand{\pathvqa}{{PathVQA}}
\newcommand{\less}{{LESS}}
\newcommand{\vpt}{{VPT}}
\newcommand{\infoprompt}{{InfoPrompt}}

\newcommand{\llm}{{Large Language Model}}
\newcommand{\llmabbrv}{{LLM}}
\newcommand{\mllm}{{Multimodal Large Language Model}}
\newcommand{\mllmabbrv}{{MLLM}}

\newcommand{\selft}{{Selective Tuning}}
\newcommand{\iterselft}{{Iterative Selective Tuning}}
\newcommand{\postselft}{{Posterior Selective Tuning}}

\newcommand{\topselft}{{Top Layer Selective Tuning}}
\newcommand{\topselftabbrv}{{Top-ST}}

\newcommand{\lastselft}{{Last Layer Selective Tuning}}
\newcommand{\lastselftabbrv}{{Last-ST}}

\newcommand{\fullselft}{{Full Layer Selective Tuning}}
\newcommand{\fullselftabbrv}{{Full-ST}}

\newcommand{\addft}{{Additive Tuning}}

\newcommand{\repft}{{Reparameterization Tuning}}
\newcommand{\strurepft}{{Structure Reparameterization Tuning}}
\newcommand{\calirepft}{{Calibration Reparameterization Tuning}}

\newcommand{\prompttuning}{{Prompt Tuning}}
\newcommand{\textualprompttuning}{{Textual Prompt Tuning}}

\newcommand{\visualprompttuning}{{Visual Prompt Tuning}}

\newcommand{\dualprompttuning}{{Dual Prompt Tuning}}

\newcommand{\adaptertuning}{{Adapter Tuning}}

\newcommand{\zeroshot}{{Zero-Shot}}
\newcommand{\fullftabbr}{{Full-FT}}

\IEEEraisesectionheading{
\section{Introduction}\label{sec:introduction}}

\begin{figure*}[t]
\centering
\includegraphics[width=\linewidth]{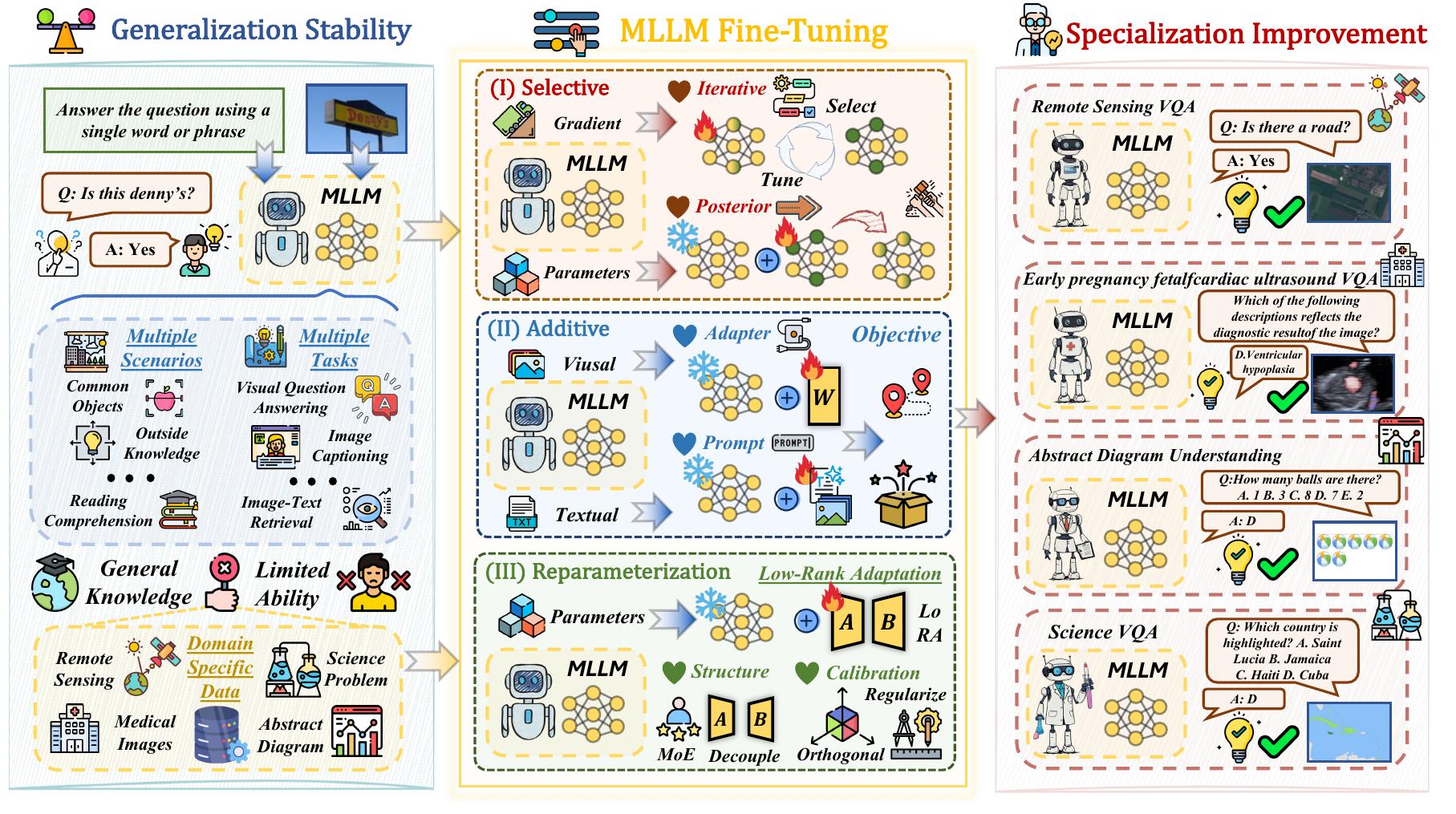}
\vspace{-20pt}
\caption{
{\color{black}
\textbf{Overview of the survey}. Best viewed in color.
}
}  
\label{fig:framework}
\vspace{-10pt}
\end{figure*}

\IEEEPARstart{W}{itness} the success of \llm{} (\llmabbrv) has remarkably transformed in the artificial intelligence landscape, demonstrating unprecedented capabilities in natural language understanding and generation {\cite{BERT_arXiv18,GPT2_OPENAI19,GPT3_NeurIPS20,PaLM_arXiv22,GPT4_arXiv23}}.
Their versatility and scalability have set new benchmarks across various domains, from conversational agents to complex problem-solving tasks.
To further enhance the applicability of \llmabbrv{}, many efforts have been made to extend \llmabbrv{} to \mllm{} (\mllmabbrv{}), which have demonstrated remarkable capabilities in generating coherent and contextually relevant descriptions from visual inputs {\cite{LLaVA_NeurIPS23,InstructBLIP_NeurIPS23,VILA_CVPR24,LLaVA15_CVPR24,VisualKnowledgeInBigModel_arXiv24}}. This fusion has expanded the horizons of AI by enabling multi-modal comprehension and interaction. \mllmabbrv{} has rapidly evolved into different complicated tasks, such as image captioning and visual question answering, to even sophisticated frameworks capable of complex reasoning and creative generation. Considering that \mllm{} is optimized on huge-scale and various-type multimodality instruction-following datasets {\cite{COCO_ECCV14,TextVQA_CVPR19,OCRVQA_ICDAR19,GQA_CVPR19,CoCoOp_CVPR22}}, it brings the powerful generalization ability on different related tasks under the open-world challenges. The advancements in \mllmabbrv{} have unlocked their potential across a wide array of applications, including autonomous driving {\cite{DriveMLLM_arXiv24,SurveyofWordModelsforAutoDriving_arXiv25}}, healthcare diagnostics {\cite{VLMforMIA_CHASE23,MedicalLLMforFuture_NPJ23}}, and remote sense {\cite{HyperSIGMA_arXiv24}}. 

Despite these compelling incentives, \mllmabbrv{} performs poorly on certain areas of expertise or private datasets {\cite{SpeandGenonCF_arXiv23, CFforLLMinCFT_arXiv23, CFinMLLMFT_CPAL24, UnifiedDelta_arXiv24, TwinMerging_NeurIPS24}}. As a result, \textit{tuning the \mllmabbrv{} for downstream tasks has emerged as an effective solution}. During the tuning stage, \mllmabbrv{} enhance task performance or align the model behavior with human expectations {\cite{PEFTMLLM_ACL24,PEFTSurvey_arXiv24}}. Despite the potential benefits of tuning, \mllmabbrv{} models often struggle to maintain satisfactory generalization ability. This is primarily due to the fact that downstream datasets often exhibit distributional divergence from the general pattern. Consequently, encouraging \mllmabbrv{} to adapt to the target distribution may lead to the tuned model losing the generality it acquired during the pre-training phase. Besides, the detrimental effect of new learning on previously acquired generic knowledge, known as catastrophic forgetting, is also a well-documented challenge for downstream adaption \cite{Connectionist_PR90,CataInter_PLM89,Catastrophic_99,CFforLLMinCFT_arXiv23,CFinMLLMFT_CPAL24}. To underscore the motivation behind our survey, we formally highlight two crucial challenges within the \mllmabbrv{} tuning field:

\noindent$\spadesuit$~\textbf{Task-Expert Specialization}. When the downstream  dataset
appear heterologous distribution behavior, pre-trained \mllmabbrv{} model appears the constrained downstream performance, thus tuning \mllmabbrv{} on the downstream tasks acts as a crucial character to become domain-expert.

\noindent$\clubsuit$~\textbf{Open-World Stabilization}. After optimization on the downstream distribution, the tuned \mllmabbrv{} model may experience catastrophic forgetting, leading to the loss of general knowledge acquired during pre-training and ultimately compromising its overall generalization capability.

In response to above obstacles, various advanced tuning strategies have been continuously developed and studied in recent years and could be broadly categorized into the following streams: \hypertarget{M1}{\textbf{\uppercase\expandafter{\romannumeral1})}}~{\color{mydarkdarkred}\textbf{\selft{}}} (\cref{sec:sft}). Focus on selecting a subset of downstream-relevant parameter elements. \hypertarget{M2}{\textbf{\uppercase\expandafter{\romannumeral2})}}~{\color{mydarkdarkblue}\textbf{\addft{}}} (\cref{sec:addft}). Add the  additional trainable modules in either input space of  inner architecture. \hypertarget{M3}{\textbf{\uppercase\expandafter{\romannumeral3})}}~{\color{mydarkdarkgreen}\textbf{\repft{}{}}} (\cref{sec:repft}). Utilize the Low-Rank Matrix technique to decomposing original parameters weight. Although existing tuning methods have been extensively studied, current \mllm{} research lacks a standardized evaluation analysis to assess the effectiveness and uniqueness of tuning strategies in \mllmabbrv{}. Moreover, the absence of systematic tuning principles introduces ambiguity in implementation workflows, resulting in redundant hyper-parameter experimentation and inefficient resource allocation. Therefore, establishing a comprehensive evaluation framework and rigorous tuning guidelines is crucial for accelerating deployment in real-world applications where time and labor constraints are critical, \eg, medical imaging analysis and remote sensing. We provide a overview  in \cref{fig:framework}.

\subsection{Prior Surveys}
\label{sec:priorsurveys}
As \mllm{} (\mllmabbrv{}) research has become a prominent research field in recent years, a large amount \mllmabbrv{}  survey papers have emerged. Existing surveys can generally be classified into two categories. 
The first category focuses on the general development of \mllm{} (\mllmabbrv{}), highlighting its applications across multiple fields. However, focusing on the conceptual framework and macro guidance would neglect in-depth exploration of specific downstream challenges and problems. The second category provides broad guidance on current tuning methods but lacks a conceptual framework and in-depth evaluations of specific tuning techniques. Although a few works {\cite{ForgettingContinual_PAMI24,ClassIncrmentalSurvey_TPAMI24}} discuss stabilization, they focus primarily on the continual learning, which investigate the neural network continue learn novel knowledge and try to maintain original knowledge {\cite{iCARL_CVPR17,EWC_PNAS17,Dark_NeurIPS20,PASS_CVPR21,DualPrompt_ECCV22,OrCo_CVPR24}} and fails to adapt into the \mllmabbrv{} field, which not only owns the unique model architecture but also has various tuning selection.  All in all, with the rapid advance of this ﬁeld, \textbf{Specialization} and \textbf{Stabilization} have been crucial aspects in \textit{tuning \mllm{} for downstream tasks}. Specialization ensures the \mllmabbrv{} performance on target distribution. Stabilization guarantees the \mllmabbrv{} to adapt to widely general tasks. Although there is a huge body of new literature, most existing surveys focus on the \textbf{narrow view} with \textbf{fragmented results}. In contrast, we argue that these two pieces interact with each other to jointly measure the practical \mllmabbrv{} deployment and this is the first work to  \textbf{simultaneously investigate} the related research development and \textbf{uniformly benchmark} multi-view experimental analysis on the downstream specialization and upstream stabilization realms.

\subsection{Structure}
A summary of the structure of this paper can be found in \textbf{Fig. 1}, which is presented as follows: 
\cref{sec:introduction} introduces the popularity of \mllm{} (\mllmabbrv{}) and outlines the technical challenges for \mllmabbrv{} tuning in the real-world scenario: Task-Expert Specialization and Open-World Stabilization.
\cref{sec:background} provides a comprehensive background on the formulation of \mllmabbrv{} and the detailed illustration tuning process. Besides, further claim the incoming challenges: specialization improvement and stabilization forgetting.
\cref{sec:tuning_taxonomy} details the taxonomy of methods: \cref{sec:sft} delves into \selft{} methods which considers selecting a partial existing parameters for tuning towards the downstream distribution. \cref{sec:addft} discusses \addft{} solution, which involves adding external parameters for adapting target domain. \cref{sec:repft} investigates the \repft{} to reconstruct existing parameter space, \ie, \lora{} module. \cref{sec:banchmark} conducts the benchmark analysis \mllmabbrv{} tuning scenario. 
\cref{sec:setup} illustrates the experiment setup with  dataset descriptions and evaluation metrics. \cref{sec:exp_com} compares different tuning method on multiple downstream datasets to discuss the impact of these technologies across diverse metrics. \cref{sec:tuning_principe} concludes the tuning principles and reveals the underlying rationale. \cref{sec:outlook} explores open challenges, potential research avenues, and promising directions for further innovation in \mllmabbrv{} tuning technologies in \cref{sec:future_direction} \cref{sec:conclusion} concludes the survey and summarizes key findings, reiterating the importance of tuning \mllmabbrv{} in real-world.

\subsection{Contribution}
\label{sec:contribution}
To fill this gap, we provide a comprehensive and timely overview that examines \textit{how specialization and stabilization behaviors emerge during \mllmabbrv{} tuning}. This paper makes the following contributions:
\begin{fullitemize}


\item \protect\includegraphics[scale=0.03,valign=c]{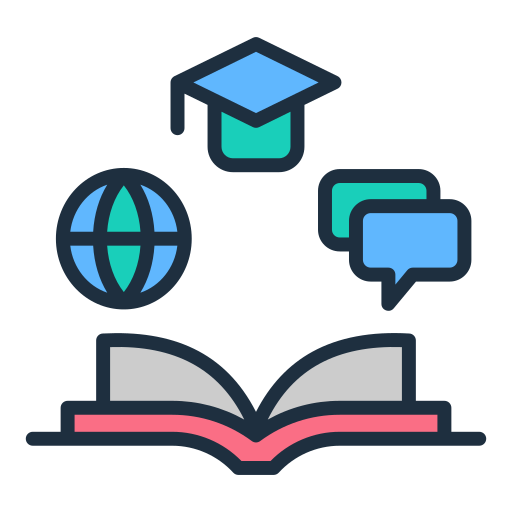} \textbf{Comprehensive Review}. We offer an in-depth exploration of specialization and stabilization during \mllmabbrv{} tuning, presenting the first state-of-the-art and systematic survey of \mllm{} tuning, covering hundreds of papers in this rapidly growing field.

\item \protect\includegraphics[scale=0.3,valign=c]{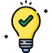} \textbf{Insightful Analysis}. We select influential tuning methods published in prestigious journals and conferences, classifying existing \mllmabbrv{} tuning.  Except for the taxonomies, an in-depth analysis of the pros and cons of these methods is also provided.

\item \protect\includegraphics[scale=0.3,valign=c]{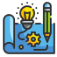} \textbf{Thorough Benchmark}. We perform an extensive benchmark analysis across various downstream scenarios with different tuning solutions. Using a set of evaluation metrics for specialization and stabilization performance, we comprehensively assess the methods effectiveness, which will provide the readers with useful guidance to select the baselines for their research.

\item \protect\includegraphics[scale=0.25,valign=c]{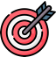} \textbf{Potential Opportunities}. We discuss future research directions that will help the community rethink and improve current designs for \mllm{} tuning in practical settings while promoting the further development of this field.
\end{fullitemize}

\section{Background}
\label{sec:background}

\subsection{History and Terminology}
\label{sec:historyandteriminology}
\subsubsection{\mllm{}}
\label{sec:mlllm_background}
The development of \llm{} (\llmabbrv{}) has revolutionized artificial intelligence, transforming the way machines understand and generate human language. Prominent examples of \llmabbrv{} include the GPT series {\cite{GPT2_OPENAI19, GPT3_NeurIPS20, GPT4_arXiv23}}, Meta LLaMA {\cite{LLaMA_arXiv23}}, and Google PaLM {\cite{PaLM_arXiv22,PaLMv2_arXiv23}}, all of which have demonstrated impressive capabilities in natural language understanding and generation. These advances have sparked significant interest in extending \llmabbrv{} to handle multi-modal inputs, particularly by incorporating vision components, leading to the development of \mllm{} (\mllmabbrv{}).
Notable \mllmabbrv{} works include Flamingo {\cite{Flamingo_NeurIPS22}}, BLIP-2 {\cite{BLIPv2_ICML23}}, \instructblip{} {\cite{InstructBLIP_NeurIPS23}}, \qwenvl{} {\cite{QwenVL_arXiv23}}, LLaVA {\cite{LLaVA_NeurIPS23,LLaVA15_CVPR24}}, and VILA {\cite{VILA_CVPR24}}, among others. These models have significantly advanced the field by enabling the joint processing of textual and visual inputs. Typically, MLLMs use a visual encoder, such as ViT {\cite{ViT_ICLR21}} or CLIP {\cite{CLIP_ICML21}}, to extract visual features, which are then projected into the word embedding space of the LLM via a connector module {\cite{LLaVA_NeurIPS23,LLaVA15_CVPR24}}, effectively treating visual input as a “foreign language” {\cite{BEiT3_CVPR23}}. The visual and textual tokens are concatenated and processed by the LLM in an auto-regressive manner to perform a wide range of vision-language tasks. For instance, LLaVA {\cite{LLaVA_NeurIPS23}} employs a linear projection layer to bridge the visual encoder and the LLM, enabling effective vision-language interaction. Therefore, the integration of vision and language in \mllmabbrv{} has opened up new possibilities across a range of general application domains, \eg, image captioning {\cite{COCO_ECCV14,Flickr_TACL14}}, where descriptive text is generated for images, and visual question answering (VQA) {\cite{OKVQA_CVPR19,GQA_CVPR19,MME_ICLR25,POPE_EMNLP23,TextVQA_CVPR19,OCRVQA_ICDAR19}}, where the model selects the correct answer based on image content. Despite their impressive capabilities, \mllmabbrv{} appears the limitations in specialized real-world applications such as Auto-Driving {\cite{DriveMLLM_arXiv24}}, Remote Sense {\cite{DLforRS_TGRS16,HyperSIGMA_arXiv24}}, and Medical Diagnosis {\cite{PathVLMforCancerDiag_arXiv24,VersatileLLMforMed_NPJ25}}. To address these challenges, tuning these models for specific tasks has emerged as a promising approach to enhance performance, improving their ability to handle task-specific requirements.

\subsubsection{Catastrophic Forgetting in \mllmabbrv{} Tuning}
Forgetting is a widely discussed issue across various research fields, such as incremental learning {\cite{ClassIncrmentalSurvey_TPAMI24,CLSurvey_TPAMI21,CLTheoryMethodApp_TPAMI24}}, federated learning {\cite{FLSurveyandBenchmarkforGenRobFair_PAMI24,FCCLPlus_PAMI23,FCCL_CVPR22}}, and test-time adaptation {\cite{OnlineTTASurvey_IJCV24,TTASurvey_IJCV25}}.  
To be precise, commonly optimized on downstream tasks {\cite{CE_AOR05}}, deep neural network is empirically proved to suffer from the \textit{catastrophic forgetting} problem {\cite{Connectionist_PR90,CataInter_PLM89,Catastrophic_99,CFforLLMinCFT_arXiv23,EWC_PNAS17,CFinMLLMFT_CPAL24}}, a significant issue where models forget previously learned information when exposed to new data. 
With respect to the \mllmabbrv{}, it brings the catastrophic forgetting of generic knowledge, which severely impairs the model transferability across previously learned datasets. Therefore, balancing the ability to fit downstream tasks while maintaining generalization becomes a crucial challenge for \mllm{}.

\subsection{Problem Formulation}
\label{sec:formulation}

\subsubsection{Training Pipeline}

\noindent In the \mllm{} (\mllmabbrv{}) architecture, the model $\theta$ typically consists of three components: the visual encoder $f$, such as \vit{} \cite{ViT_ICLR21}, the \llmabbrv{} module $g$, exemplified by \vicuna{} \cite{Vicuna_arXiv23} and \llama{} \cite{LLaMA_arXiv23}, and the connector module $\varphi$ \cite{LLaVA_NeurIPS23,InstructBLIP_NeurIPS23,LLaVA15_CVPR24,VILA_CVPR24}. Generally , \mllmabbrv{} follows the paradigm to fuse the pre-trained vision encoder {\cite{CLIP_ICML21,ViT_ICLR21}} into the representation space of the  \llm{}, \eg, \llama{} {\cite{LLaMA_arXiv23}} and \vicuna{} {\cite{Vicuna_arXiv23}}, via the connector module   {\cite{InstructBLIP_NeurIPS23,LLaVA_NeurIPS23,LLaVAHR_arXiv24}}. To be detailed, given a query instance, the input comprises both a visual image $x^v$ and a textual instruction $x^t$, with the corresponding label being a language response $y$. First, the visual features are extracted as $z^v = f(x^v)$, and then the trainable projection $\varphi$ is applied to map $z^v$ into language embedding tokens, $h^v = \varphi \cdot z^v$. The textual token is similarly generated as $h^t = \text{Tokenize}(x^t)$. These visual and textual tokens are then concatenated and passed through the \llmabbrv{} module $g$ to generate the language output, $\mathbf{y} = \delta([h^v, h^t])$. 
With respect to \mllmabbrv{} tuning process, its \llmabbrv{} module often contains hundreds of billions of parameters. Therefore, the parameter-efficient tuning paves an feasible solution to tuning the \mllmabbrv{} via few trainable parameters. In our work, following previous \mllmabbrv{} tuning approaches and benchmarks \cite{PEFTMLLM_ACL24,ModelTailor_ICML24,SPIDER_arXiv24}, we select and tune a subset of the trainable parameters, denoted as $w$, to adapt to the downstream task $\mathcal{T}$ with distribution $D^\mathcal{T}$. The typically learnable modules include the connector module $\varphi$ and selected layers in the \llmabbrv{} block $\delta$, where $W = \{\varphi, \delta\}$. This default \mllmabbrv{} optimization procedure is defined as follows:
\begin{equation}
\setlength\abovedisplayskip{2pt} \setlength\belowdisplayskip{2pt}
\arg \min_{W} \mathbb{E}_{(x^v,x^t,y) \in {\mathcal{D}^\mathcal{T}}} \mathcal{L}\left(\delta([\varphi(h^v),h^t]), y \right).
\label{eq:ce_loss}
\end{equation}
Consequently, it directly adapts to the target samples to enhance specialization while compromising the stabilization of learned knowledge, as the model pay no attention to  previous distributions. We further provide the training description in 

\begin{figure}[t]
\begin{center}
\includegraphics[width=\linewidth]{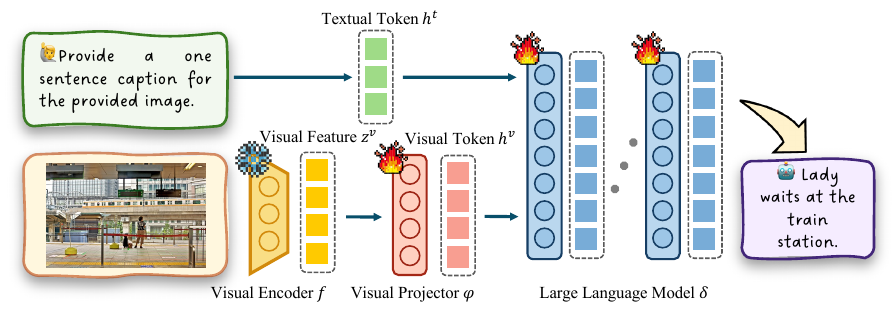}
\end{center}
\vspace{-15pt}
\caption{
\textbf{The flow chart of \mllm{} Tuning Paradigm}. Refer to \cref{sec:formulation} for details. 
}
\vspace{-10pt}
\label{fig:flowchart}
\end{figure}

\subsubsection{Challenges Declaration}
To define key aspects in evaluating \mllm{} tuning, we introduce the following concepts:
\begin{definition}[Specialization Improvement] \textit{Let $\mathcal{T}$ denote a downstream task, and $\mathcal{D}^\mathcal{T}$ represent the associated  dataset. Consider a pre-trained model $\theta$ and a tuned model $\theta^*$. Define the performance function $\phi(\theta, \mathcal{D}^\mathcal{T})$, which evaluates the performance of model $\theta$ on the downstream task $\mathcal{T}$ using the dataset $\mathcal{D}^\mathcal{T}$.
The following declaration holds:
\begin{equation} \exists \; \theta^* , \; \operatorname{s.t.} \; \phi(\theta^*, \mathcal{D}^\mathcal{T}) \geq \phi(\theta, \mathcal{D}^\mathcal{T}). \end{equation}
The object of this study is to maximize the improvement in performance between the $\theta^*$ and the $\theta$ on the downstream task:
\begin{equation} {\arg \max}_{\theta^*}  \left \{ \phi(\theta^*, \mathcal{D}^\mathcal{T}) - \phi 
(\theta, \mathcal{D}^\mathcal{T}) \right \}, \end{equation}
where the goal is to find the model $\theta^*$ that maximizes the specialization improvement over the pre-trained model $\theta$ on the task $\mathcal{T}$.}
\end{definition}

\begin{definition}[Stabilization Forgetting] 
\textit{Let $\mathcal{S}$ denote the source distribution from pre-training, and $\mathcal{D}^\mathcal{S}$ represent the corresponding dataset. Consider a pre-trained model $\theta$ and a tuned model $\theta^*$. Define the performance function $\phi(\theta, \mathcal{D}^\mathcal{S})$, which evaluates the performance of model $\theta$ on the source task using the dataset $\mathcal{D}^\mathcal{S}$.
The following circumstance may hold:
\begin{equation} 
\exists \; \theta^* \; \operatorname{s.t.} \; \phi(\theta^*, \mathcal{D}^\mathcal{S}) \leq \phi(\theta, \mathcal{D}^\mathcal{S}). 
\end{equation}
This indicates that after tuning, the model $\theta^*$ may experience a degradation in performance on the original source task.
The objective of this study is to minimize the absolute difference between the performance of $\theta^*$ on the source dataset $\mathcal{D}^\mathcal{S}$ and the performance of the pre-trained model $\theta$ on the downstream task $\mathcal{T}$: 
\begin{equation}  
{\arg \min}_{\theta^*} \; |\phi(\theta^*, \mathcal{D}^\mathcal{S}) - \phi(\theta, \mathcal{D}^\mathcal{S})|.
\end{equation} 
The goal is to ensure that the performance drop due to forgetting is controlled while maintaining effective adaptation to the downstream task $\mathcal{T}$. This stabilization ensures that the model retains as much knowledge as possible from pre-training while still improving on the new task.}
\end{definition}

We further provide a notations summary in \cref{tab:notation} to help readers quickly understand  key terms used in our work.

\begin{table}[t]
\caption{\small{
\textbf{Notations} table.
}}
\label{tab:notation}
\vspace{-10pt}
\centering
\resizebox{\linewidth}{!}{
\setlength\tabcolsep{4pt}
\renewcommand\arraystretch{1.2}
\begin{tabular}{clIcl}
\hline\thickhline
\rowcolor{mydarkyellow}
 & Description &  & Description
 \\ \hline\hline
$\theta$ & \mllmabbrv{} &
$f$ & Vision encoder \\
$\varphi$ & Connector  &
$\delta$ & \llm{} \\
$x^v$ & Visual image  &  
$x^t$ & Textual instruction  \\
$z^v$ & Visual feature  &
$h^v$ & Visual token  embedding \\
$h^t$ & Text token embedding &
$z$ & Logits output \\
$\mathbf{y}$ & Language response & 
$y$ & Ground truth  \\
$\mathcal{S}$ & Source dataset &
$\mathcal{T}$ & Target dataset \\
$\mathcal{D}$ & Data distribution &
$\mathcal{L}$ & Loss function \\
$W$ & Learnable parameter  &
$\phi$ & Performance function \\
$\mathcal{M}$ & Updating mask &
$\odot$ & Hadamard product \\
$\eta$ & Learning rate &
$g$ & Gradient \\
$\mathcal{E}$ & Specialization improvement &
$\mathcal{F}$ & Stabilization forgetting \\
$\mathcal{A}$ & Accuracy function &
$\mathcal{O}$ & O-Average metric \\
\end{tabular}%
}
\vspace{-5pt}
\end{table}

\begin{table}[t]\small
\caption{\textbf{Summary of essential characteristics for reviewed solutions in \selft{} (\cref{sec:sft})}. Para. and Grad. denotes the parameter and gradient signals.}
\label{tab:sft_summary}
\vspace{-10pt}
\centering
{
\resizebox{\columnwidth}{!}{
\setlength\tabcolsep{2pt}
\renewcommand\arraystretch{1.2}
\begin{tabular}{rIc|c|cc}
\hline\thickhline
\rowcolor{mydarkred}
\textbf{Methods} & \textbf{Venue} & \textbf{Importance Criteria} & \textbf{Para.} & \textbf{Grad.}  \\
\hline\hline
\multicolumn{4}{l}{\textit{\textbf{\iterselft{}}}} \\  
\hline
\rowcolor{mylightred}
\magpurne{} \cite{MagnitudePurning_NeurIPS15} & \pub{NeurIPS'15} & Weight Magnitude  & {\color{mydarkdarkred2}\faCheck} &\\
\child{} \cite{CHILDTUNING_EMNLP21} & \pub{EMNLP'21} & Fisher information & & {\color{mydarkdarkred2}\faCheck} \\
\rowcolor{mylightred}
\pst{} \cite{PST_IJCAI22} & \pub{IJCAI'22} & Rank decomposition & {\color{mydarkdarkred2}\faCheck} & {\color{mydarkdarkred2}\faCheck}  \\
\ltsft{} \cite{LTSFT_ACL22} & \pub{ACL'22} & Lottery ticket hypothesis & {\color{mydarkdarkred2}\faCheck}  \\
\rowcolor{mylightred}
\rose{} \cite{ROSE_arXiv22} &  \pub{arXiv'22} & Adversarial perturbation &  & {\color{mydarkdarkred2}\faCheck}\\
\losparse \cite{LoSparse_ICML23} & \pub{ICML'23} & Low-rank and sparse & {\color{mydarkdarkred2}\faCheck}   & \\
\rowcolor{mylightred}
\fishdip{} \cite{FISHDIP_EMNLP23} & \pub{EMNLP'23} & Fisher information & & {\color{mydarkdarkred2}\faCheck} \\
\gps{}  \cite{GPS_CVPR24} & \pub{CVPR'24} & Gradient modulus & & {\color{mydarkdarkred2}\faCheck} \\
\rowcolor{mylightred}
\spu{} \cite{SPU_CVPR24} & \pub{CVPR'24} & Gradient magnitude   & & {\color{mydarkdarkred2}\faCheck} \\
\sift{} \cite{SIFT_ICML24} & \pub{ICML'24} & Absolute gradient value & & {\color{mydarkdarkred2}\faCheck}  \\
\rowcolor{mylightred}
\spider{} \cite{SPIDER_arXiv24}  & \pub{arXiv'24} & Importance discrepancy &{\color{mydarkdarkred2}\faCheck}   & {\color{mydarkdarkred2}\faCheck} \\
\alphaedit{} \cite{AlphaEdit_ICLR25}  & \pub{ICLR'25} & Singular Value Decompose &{\color{mydarkdarkred2}\faCheck}  &\\ 
\hline\hline
\multicolumn{4}{l}{\textit{\textbf{\postselft{}}}} \\  
\hline
\rowcolor{mylightred}
FisherMerge\cite{FisherAveraging_NeurIPS22} & \pub{NeurIPS'22} & Fisher Information  &{\color{mydarkdarkred2}\faCheck} & \\
\ties{} \cite{TIES_NeurIPS23} & \pub{NeurIPS'23} & Trim, elect and pick & {\color{mydarkdarkred2}\faCheck} &\\
\rowcolor{mylightred}
\dare{} \cite{DARE_ICML24}  & \pub{ICML'24} & Bernoulli selection &  &\\
\twinmerge{} \cite{TwinMerging_NeurIPS24} & \pub{NeurIPS'24} & SVD and MoE & {\color{mydarkdarkred2}\faCheck} & \\
\rowcolor{mylightred}
\tailor{} \cite{ModelTailor_ICML24}  & \pub{ICML'24} & Hessian Matrix & {\color{mydarkdarkred2}\faCheck} &\\
\cart{} \cite{CART_arXiv24} & \pub{arXiv'24} & Low-rank approx &{\color{mydarkdarkred2}\faCheck}  &\\
\rowcolor{mylightred}
\pcb{} \cite{PCB_NeurIPS24} & \pub{NeurIPS'24} & Intra and Inter Balance &{\color{mydarkdarkred2}\faCheck} & \\
\emrmerging{} \cite{EMRMerging_NeurIPS24} & \pub{NeurIPS'24} & Task shared and specifc &{\color{mydarkdarkred2}\faCheck} & \\
\hline\thickhline
\end{tabular}}}
\end{table}
\section{Tuning Taxonomy}
\label{sec:tuning_taxonomy}

\subsection{\selft{}}
\label{sec:sft}
\selft{} focuses on tuning a subset of downstream-relevant parameters, rather than the entire model architecture. The rationale behind is that not all parameters equally contributes to the target distribution {\cite{Pruning_ICLR17,LTH_ICLR19,RethinkingNetPruning_ICLR19,FISH_NeurIPS21,Chase_NeurIPS23,Wanda_ICLR24}}. Thus, it is feasible to select and optimize the candidate elements that are crucial for downstream behaviors. Thus, construing the updating mask $\mathcal{M}$ act as the kernel character within this paradigm and could be broadly categorized into two major types: \iterselft{} and \postselft{}.

\subsubsection{\iterselft{}}
\label{sec:iterselft}
\iterselft{} \cite{PST_IJCAI22,LTSFT_ACL22,LoSparse_ICML23,DARE_ICML24,GPS_CVPR24,SPU_CVPR24} focuses on localizing and updating target elements during the training process. We could derive the following methods description.
\begin{equation}
\setlength\abovedisplayskip{2pt} \setlength\belowdisplayskip{2pt}
\begin{aligned}
g &=  \nabla \mathcal{L}\left(\delta([\varphi(h^v),h^t]), y \right), \\
W &= W  - \mathcal{M} \odot \eta  g,
\end{aligned}
\label{eq:update}
\end{equation}
where $g=\mathcal{L}\left(\delta([\varphi(h^v),h^t]), y \right)$ is the gradient of the cross-entropy loss with respect to $W$. $\eta$ denotes the learning rate. $\mathcal{M}$ means the updating mask.  As a result, the gradients of unselected parameters are zeroed out and excluded from updates. Relevant methods typically combine the original parameter weights, denoted as $W^*$, with the current gradients, $g$, to construct the partial update mask and could be further divided into the following types:

\noindent$\bullet$~\textbf{Pretrained Magnitude Guidance.} The pre-trained weights are used as a data-free criterion to compute the importance of weights based solely on their magnitudes, without involving any data \cite{MagnitudePurning_NeurIPS15}. The underlying assumption is that weights with larger magnitudes carry higher importance for previously acquired knowledge \cite{MagnitudePurning_NeurIPS15, LTH_ICLR19, Wanda_ICLR24, FedCPA_ICCV23, DLTH_ICLR22}. Consequently, a range of methods has been developed to construct binary masks that freeze the associated upstream elements, thereby preserving the general task ability and mitigating catastrophic forgetting. For instance, \ltsft{} \cite{LTSFT_ACL22} adopts the Lottery Ticket Hypothesis \cite{LTH_ICLR19, DLTH_ICLR22} to build task-specific masks. Similarly, \alphaedit{} \cite{AlphaEdit_ICLR25} employs singular value decomposition to project parameter perturbations onto the null space. \losparse{} \cite{LoSparse_ICML23} approximates the weight matrix by the
sum of a low-rank matrix and a sparse matrix.

\noindent$\bullet$~\textbf{Current Gradient Information.} During the backward pass, gradients typically reflect the rate of change of the output with respect to the model parameters, providing intensity information about the learning signal imposed on each parameter element for the optimization objective \cite{Fisher_1922, NatureGradient_arXiv13, EWC_PNAS17, Snip_ICLR19, StableCL_NeurIPS20, MovementPruning_NeurIPS20, SIFT_ICML24}. Therefore, utilizing gradients and their variants serves as a feasible signal for discovering downstream critical elements. For example, \child{} \cite{CHiLS_ICML23} and \fishdip{} consider the Hessian matrix, which is the second derivative of the gradient, to construct downstream-relevance masks. \gps{} \cite{GPS_CVPR24} and \spu{} \cite{SPU_CVPR24} both leverage the momentum gradient magnitude.

\noindent$\bullet$~\textbf{Multi Knowledge Collaboration.} Recent works have explored the simultaneous use of both prior (pre-trained) and posterior (current) knowledge to construct candidate tuning masks, enhancing model performance in downstream tasks. For example, \pst{} \cite{PST_IJCAI22} leverages the low-rank structure of both the weights and gradients to construct a data-driven importance score mask, capitalizing on the structure of these elements to guide the optimization process. Recently, \spider{} \cite{SPIDER_arXiv24} combines pre-trained weights with ongoing gradients to measure discrepancies in parameter importance, enabling more precise strategies for parameter update allocation. This paradigm facilitates efficient tuning while preserving previously acquired knowledge by utilizing both historical and real-time information.

\subsubsection{\postselft{}}
\postselft{}~\cite{DARE_ICML24,ModelTailor_ICML24,TwinMerging_NeurIPS24,PCB_NeurIPS24} operates on the tuned model and aims to separately maintain the partially important downstream elements, while the remaining parts are set to the original values. Thus, this paradigm could be regard as merging the trained model into the original ones. Existing explorations mainly concentrate on 
\begin{equation}
\setlength\abovedisplayskip{2pt} \setlength\belowdisplayskip{2pt}
W=  W^* \odot (1-\mathcal{M}) +  W  \odot \mathcal{M} \\
\label{eq:maskupdate}
\end{equation}
Motivated by the above formulation, existing literature could be categorized into the following two groups:

\noindent$\bullet$~\textbf{Task Specialization Merge.} This paradigm is driven by a straightforward fusion strategy to derive a sparse mask that identifies task-specialized model updates. Early works, such as \dare{} \cite{DARE_ICML24}, directly preserve partial parameter updates in a random manner. Recently, a growing body of research has employed various metrics to evaluate the critical task elements. For example, FisherMerge \cite{FisherAveraging_NeurIPS22} identify the importance of individual parameters using Fisher information matrix {\cite{TutorialFisher_JMP,Fisher_1922,Fishr_ICML22,SDFC_ECCV24}} and uses it to measure parameter importance.
\tailor{} \cite{ModelTailor_ICML24} conducts salience and sensitivity analysis to select the candidate sparse specialization mask \cite{SparseGPT_ICML23}.

\noindent$\bullet$~\textbf{Task Collaboration Merge.} Facing multiple tuned \mllmabbrv{} fusion condition, this direction focus on to alleviate the task interface and preserve multi-party information. For instance, both \cart{} \cite{CART_arXiv24} and \twinmerge{} {\cite{TwinMerging_NeurIPS24}} utilize the Singular Value Decomposition to excavate the task-relevant knowledge. \twinmerge{} further utilize the Mixture of Experts {\cite{MoEforMulitClass_NeurNet99,MoE_AIR14,EvaluateMoE_NeurIPS90}}. Recently, \pcb{} \cite{PCB_NeurIPS24} considers the grained level parameter importance to utilize the intra-balancing to gauge parameter significance within individual tasks and inter-balancing to assess parameter similarities across different tasks. We provide a review for relative methods in \cref{tab:sft_summary}.

\subsubsection{Pros and Cons}
With respect to the \selft{} paradigm, this approach offers several advantages, as detailed below:
\begin{itemize}
\item[\faThumbsOUp] \textbf{Structural Compatibility.} The \selft{} paradigm seamlessly integrates with diverse architectures by optimizing only a subset of parameters relevant to the target task. This enables efficient adaptation to pre-trained models without requiring substantial architectural modifications.

\item[\faThumbsOUp] \textbf{Inference Efficiency.} As \selft{} selectively tunes only a subset of the original architecture’s parameters, it effectively manages model complexity while maintaining inference efficiency across diverse downstream tasks.
\end{itemize}
Despite these advantages, there are also some limitations associated with \selft{}:
\begin{itemize}
\item[\faThumbsODown] \textbf{Pre-defined Mask Ratio.} The success of \selft{} relies on the pre-defined mask ratio, which determines how many parameters are tuned. An improper ratio can lead to suboptimal performance by either retaining irrelevant parameters or ignoring important ones.

\item[\faThumbsODown] \textbf{Incremental Memory Cost.} While \selft{} tunes only a subset of parameters, maintaining pre-trained weights and sparse update masks increases memory costs. This can be problematic in memory-constrained environments, such as edge devices.
\end{itemize}

\begin{table}[t]\small
\caption{\textbf{Summary of essential characteristics for relative solutions in \addft{} (\cref{sec:addft})}. Vis. and Tex. means the operations on visual and textual branch.}
\label{tab:addft_summary}
\vspace{-10pt}
\centering
{
\resizebox{\columnwidth}{!}{
\setlength\tabcolsep{2pt}
\renewcommand\arraystretch{1.2}
\begin{tabular}{rIc|c|cc}
\hline\thickhline
\rowcolor{mydarkblue}
\textbf{Methods} & \textbf{Venue} & \textbf{Highlight} & \textbf{Vis.} & \textbf{Tex.}  \\
\hline\hline
\multicolumn{4}{l}{\textit{\textbf{Adapter}}} \\  
\hline
\rowcolor{mylightblue}
\adapter{} \cite{Adapter_ICML19} & \pub{ICML'19} & Add a few trainable parameters  & & {\color{mydarkdarkblue2}\faCheck} \\
\lst{} \cite{LST_NeurIPS22} & \pub{NeurIPS'22} & Ladder side-tuning & {\color{mydarkdarkblue2}\faCheck}  & {\color{mydarkdarkblue2}\faCheck}    \\
\rowcolor{mylightblue}
\tipadater{} \cite{TipAdapter_ECCV22} & \pub{ECCV'22} & Key-value cache model & {\color{mydarkdarkblue2}\faCheck}  &\\
\clipadapter{} \cite{ClipAdapter_IJCV23} & \pub{IJCV'23} & Additional bottleneck layers & {\color{mydarkdarkblue2}\faCheck}  & {\color{mydarkdarkblue2}\faCheck}\\
\rowcolor{mylightblue}
\fdt{} \cite{FDT_CVPR23} & \pub{CVPR'23} &  Finite discrete tokens as anchors & {\color{mydarkdarkblue2}\faCheck}  & {\color{mydarkdarkblue2}\faCheck}\\
\san{} \cite{SAN_CVPR23} & \pub{CVPR'23} &  Mask proposals and attention biases & {\color{mydarkdarkblue2}\faCheck} &\\
\rowcolor{mylightblue}
\ape{} \cite{APE_ICCV23} & \pub{ICCV'23} &  Adaptive prior refinement & {\color{mydarkdarkblue2}\faCheck}  & {\color{mydarkdarkblue2}\faCheck} \\
TaskRec \cite{TaskRes_CVPR23} & \pub{CVPR'23} & Prior-independent residuals &  &{\color{mydarkdarkblue2}\faCheck}\\
\rowcolor{mylightblue}
\metaadapter{} \cite{MetaAdapter_NeurIPS23} & \pub{NeurIPS'23} & Residual network and meta-test &{\color{mydarkdarkblue2}\faCheck} &{\color{mydarkdarkblue2}\faCheck}\\
\graphadapter{} \cite{GraphAdapter_NeurIPS23} & \pub{NeurIPS'23} &  Dual knowledge sub-graph & {\color{mydarkdarkblue2}\faCheck} & {\color{mydarkdarkblue2}\faCheck}\\
\rowcolor{mylightblue}
\cpr{} \cite{CPR_CVPR24} & \pub{CVPR'24}  &  Conditional multimodal adapter& {\color{mydarkdarkblue2}\faCheck} & {\color{mydarkdarkblue2}\faCheck} \\

\hline\hline
\multicolumn{4}{l}{\textit{\textbf{Prompt}}} \\  
\hline
\rowcolor{mylightblue}
\autoprompt{} \cite{AutoPrompt_EMNLP20} & \pub{EMNLP'20} & Automatically-constructed prompts & & {\color{mydarkdarkblue2}\faCheck}\\
\lpaqa{} \cite{LPAQA_ACL20} & \pub{EMNLP'20} & Mining and paraphrasing generation & & {\color{mydarkdarkblue2}\faCheck}\\
\rowcolor{mylightblue}
\prefixtuning{} \cite{PrefixTuning_arXiv21} & \pub{EMNLP'21} & Optimize Continuous Prompts & & {\color{mydarkdarkblue2}\faCheck}\\
\coop{} \cite{CoOp_IJCV22} & \pub{IJCV'22} & Context optimization & & {\color{mydarkdarkblue2}\faCheck} \\
\rowcolor{mylightblue}
\vpt{} \cite{VPT_ECCV22} & \pub{ECCV'22} & Visual prompt tuning & {\color{mydarkdarkblue2}\faCheck} &  \\
\cocoop{} \cite{CoCoOp_CVPR22} & \pub{CVPR'22} & Conditional context optimization & & {\color{mydarkdarkblue2}\faCheck}  \\
\rowcolor{mylightblue}
\dualcoop{} \cite{DualCoOp_NeurIPS22} & \pub{CVPR'22} & Positive and negative contexts & & {\color{mydarkdarkblue2}\faCheck} \\
\kapt{} \cite{KAPT_ICCV23} & \pub{ICCV'23} & Category-related external knowledge &  & {\color{mydarkdarkblue2}\faCheck} \\
\rowcolor{mylightblue}
\prograd{} \cite{ProGrad_ICCV23} & \pub{ICCV'23} & Update prompt with aligned gradient & & {\color{mydarkdarkblue2}\faCheck}  \\
\plot{} \cite{PLOT_ICLR23} & \pub{ICLR'23} & Optimal transport & & {\color{mydarkdarkblue2}\faCheck}  \\
\rowcolor{mylightblue}
\kgcoop{} \cite{KgCoOp_CVPR23} & \pub{CVPR'23} & Align learnable and crafted prompt & & {\color{mydarkdarkblue2}\faCheck} \\
\promptsrc{} \cite{PromptSRC_ICCV23} & \pub{ICCV'23} & Self-regulating prompt  & {\color{mydarkdarkblue2}\faCheck} & {\color{mydarkdarkblue2}\faCheck} \\
\rowcolor{mylightblue}
\maple{} \cite{MaPLe_CVPR23} & \pub{CVPR'23} & Coupling function for mutual synergy & {\color{mydarkdarkblue2}\faCheck}  & {\color{mydarkdarkblue2}\faCheck}  \\
\dapt{} \cite{DAPT_ICCV23} & \pub{ICCV'23} & Text-Inter and vision-intra dispersion& {\color{mydarkdarkblue2}\faCheck} & {\color{mydarkdarkblue2}\faCheck}  \\
\rowcolor{mylightblue}
\promptkd{} \cite{PromptKD_CVPR24} & \pub{CVPR'24} & Student prompt distillation  & {\color{mydarkdarkblue2}\faCheck} & \\
\protext{} \cite{ProText_CVPR24} & \pub{CVPR'24} & Embed contextual knowledge from \llmabbrv & &  {\color{mydarkdarkblue2}\faCheck} \\
\rowcolor{mylightblue}
\dept{} \cite{DePT_CVPR24} & \pub{CVPR'24} & Channel adjusted transfer  & &  {\color{mydarkdarkblue2}\faCheck}\\
\argue{} \cite{ArGue_CVPR24} & \pub{CVPR'24} & Attribute-Guided Prompt Tuning & &  {\color{mydarkdarkblue2}\faCheck}\\
\rowcolor{mylightblue}
\damp{} \cite{DAMP_CVPR24} & \pub{CVPR'24} & Exploit domain-invariant semantics &  {\color{mydarkdarkblue2}\faCheck} &  {\color{mydarkdarkblue2}\faCheck} \\
\hline\thickhline
\end{tabular}}}
\end{table}

\subsection{\addft}
\label{sec:addft}
\addft{} introduces additional trainable parameters without altering the original model parameters. In contrast, existing methods typically inject learnable parameters from the input space prompt: \prompttuning{} \cref{sec:prompttuning} or the inner architecture adapter: \adaptertuning{} \cref{sec:adaptertuning}. This approach leverages these learnable parameters to optimize the model for specific tasks, thereby transferring task knowledge to downstream tasks by adjusting parameters in pre-trained \mllm{}.

\subsubsection{\prompttuning{}}
\label{sec:prompttuning}
Prompting \cite{PromptTuning_EMNLP21,AutoPrompt_EMNLP20,BERTofTheseus_EMNLP20,LPAQA_ACL20,LMBFF_ACL21,OptiPrompt_NAACL21,LPT_arXiv22,Xprompt_arXiv22,IPT_arXiv21} is a fundamental technique in Natural Language Processing (NLP), often used as a transfer approach or to provide specific instructions for downstream tasks. Incorporating a prompt module has proven to be an effective and efficient tool for calibrating model behavior. A range of studies have been proposed, focusing on injecting the prompt from three key perspectives.

\noindent$\bullet$~\textbf{\textualprompttuning.}
Prompt learning has become a widely adopted adaptation technique for \mllmabbrv{}. \coop \cite{CoOp_IJCV22} was among the first to utilize learnable context tokens to prompt the language encoder of CLIP for visual classification. In this framework, the text prompt $\bm{P}_t$ is represented as a learnable vector $v$ combined with the class token. The input to the text encoder is then expressed as:
\begin{equation}
\setlength\abovedisplayskip{2pt} \setlength\belowdisplayskip{2pt}
\Tilde{h}^t = [v_1, v_2, \cdots, v_L,\texttt{CLASS}].\footnote{We adhere to the definition in \clip{} for clarity.}
\end{equation}

However, prompting has been shown to overfit downstream data distributions, making it difficult to maintain generalization ability \cite{SubPT_TCSVT23,ProMetaR_CVPR24,DePT_CVPR24,Dept_ICLR24,NOAH_arXiv22}. Consequently, numerous efforts have been made to enhance the generalization of prompts by introducing techniques such as self-regularization calibration \cite{CoCoOp_CVPR22,DAPT_ICCV23,PromptSRC_ICCV23,ProGrad_ICCV23,KgCoOp_CVPR23,PLOT_ICLR23,DualCoOp_NeurIPS22,DualCoOpPlus_PAMI23,PTP_EMNLP23} and external contextual knowledge libraries \cite{KAPT_ICCV23,ProText_CVPR24}. For instance, \cocoop{} \cite{CoCoOp_CVPR22} learns instance-conditioned prompts through a two-layer network to inject knowledge from individual images. Similarly, \kapt{} retrieves textual descriptions of task labels from a Wikipedia knowledge base.

\noindent{} \textbf{\visualprompttuning.}
With regard to the visual prompt, we follow the approach outlined in \cite{VPT_ECCV22} by inserting a visual prompt $\bm{P}_v$ consisting of learnable vectors $u$ between the class token $\mathtt{CLS}$ and the image patch embeddings $\bm{E}$ in the image encoder. The resulting input representation is:
\begin{equation}
\setlength\abovedisplayskip{2pt} \setlength\belowdisplayskip{2pt}
\Tilde{x} = [\mathtt{CLS}, u_1, u_2, \dots, u_L, \bm{E}].\footnote{We adhere to the definition in \clip{} for clarity.}
\label{eq:visual_prompt}
\end{equation}
Empirical studies have demonstrated that visual prompts serve as an effective mechanism for adapting pretrained Vision Transformers to downstream tasks {\cite{GatedPromptTuning_ICML23,VPforLargeModel_arXiv22}}. Moreover, visual prompts have been shown to outperform full tuning, particularly in scenarios where task objectives are misaligned or data distributions exhibit significant discrepancies {\cite{LPFT_ICLR22,VPTorFinetuning_ICLR24}}.

\noindent{} \textbf{\dualprompttuning.}
Recently, several studies have explored injecting prompts into both the visual and textual modalities to simultaneously adapt the behaviors of these two branches \cite{PromptSRC_ICCV23,MaPLe_CVPR23,DAPT_ICCV23}. For example, \maple{} \cite{MaPLe_CVPR23} introduces branch-aware hierarchical prompts and utilizes a coupling function to induce mutual synergy. \promptsrc{} \cite{PromptSRC_ICCV23} proposes regularization through mutual agreement maximization and prompt self-ensembling modules. Meanwhile, \dapt{} \cite{DAPT_ICCV23} encourages both inter-dispersion of text prompts and intra-dispersion of visual prompts. 

\subsubsection{\adaptertuning{}}
\label{sec:adaptertuning}
Adapter tuning \cite{Adapter_ICML19} is a lightweight neural module that facilitates parameter-efficient tuning of pre-trained models, such as BERT \cite{BERT_arXiv18}, in natural language processing and other domains. This method introduces small, task-specific modules into a pre-trained network, allowing it to be adapted to new tasks without requiring extensive modifications to the original model weights, thus reducing computational costs and memory usage.

\noindent \textbf{Individual Behavior Adapter.} This category of methods aims to refine individual-level textual classifiers or visual feature patterns through simple yet efficient feature modulation for specific tasks, typically focused on the output side. For example, \clipadapter{} \cite{ClipAdapter_IJCV23} introduces a simple bottleneck layer to adjust the textual and visual embeddings of \mllmabbrv{}. TaskRes \cite{TaskRes_CVPR23} utilizes learnable, task-specific parameters as prior-independent residuals to update the textual embeddings, improving task-specific performance.

\noindent \textbf{Group Behavior Adapter.} Recent literature has expanded this concept to consider group behavior by incorporating downstream sample relationships. For instance, \graphadapter{} \cite{GraphAdapter_NeurIPS23} exploits explicit knowledge structures to establish correlations between different semantic representations in both textual and visual modalities. Additionally, \cpr{} \cite{CPR_CVPR24} leverages input images alongside visual and textual prototypes to capture structured knowledge that is pertinent to downstream tasks, further enhancing model adaptability. An overview for existing \addft{} exploration is plot in \cref{tab:addft_summary}.

\subsubsection{Pros and Cons}
Towards \addft{}, we discuss the relative advantages and disadvantages of this paradigm to better define its suitable scope. As for the advantages, we list as follows:
\begin{itemize}
\item[\faThumbsOUp] \textbf{Plug and Play}. \addft{} focuses on introducing trainable parameters, such as visual prompts, textual prompts, and adapter modules. This paradigm does not alter the original parameter space and effectively preserves the pre-existing knowledge.
\end{itemize}
In response to the weakness in \addft{}, we plot the following shortcomings:
\begin{itemize}
\item[\faThumbsODown] \textbf{Inference Efficiency Constraints.} As the number of additional parameters increases, it inevitably leads to a decrease in inference speed. Specifically, prompts extend the input token length, and adapters serve as essential components for forward propagation. Consequently, \addft{} results in increased inference overhead.

\item[\faThumbsODown]  \textbf{Downstream Overfitting Binding.} Regarding the prompt and adapter modules, it has been empirically observed that they tend to overfit on downstream tasks \cite{PromptSRC_ICCV23,SubPT_TCSVT23}. As a result, the effectiveness of \addft{} is strictly constrained to the current data distribution.

\item[\faThumbsODown]  \textbf{Architecture Compatibility Burden.} Specifically, the prompt module needs to be integrated into transformer blocks \cite{Attention_NeurIPS17,BERT_arXiv18,ViTGAN,LLaMA_arXiv23}. Thus, \addft{} requires a specialized architecture, limiting its compatibility with other architectures.

\end{itemize}

\subsection{\repft{}}
\label{sec:repft}
\repft{} investigates the training the parameters in low-rank matrices,
significantly reducing the number of trainable parameters, which normally called the Low-Rank Adaptation (\lora{}) {\cite{LoRA_ICLR22,AdaLoRA_arXiv23,LiGO_ICLR23,Flora_ICML24,DoRA_ICML24,LoraSurvey_FCS25}}. \lora{} is a pioneering approach for adapting large, pre-trained models to downstream tasks via efficient, low-rank updates. Let  $W^*$  be the frozen pre-trained weight matrix. \lora{} introduces two trainable matrices $A$ and $B$ of significantly lower rank (i.e.,  $r \ll \text{dim}(W)$ ), forming a parallel branch in each model layer. The effective model update is given by: $\Delta W = B A$ Thus, the updated weights are: $W = W^* + \Delta W = W^* + BA$. Because only $ A $ and $B$ are learned during optimization, the number of trainable parameters and the associated GPU memory requirements are substantially reduced. Moreover, the inclusion of these low-rank adapters as a bypass to the original model architecture—often within self-attention modules—allows \lora{} to approximate the intrinsic rank of the adaptation needed. This makes \lora{} a resource-efficient technique for quickly tuning large language models using minimal data, ideal for many practical tasks. Driven by the above formulation, follow-up works  could be dived into the \strurepft{} {\cref{sec:strurepft}} and \calirepft{} {\cref{sec:calirepft}} branches to boost the effectiveness.

\subsubsection{\strurepft}
\label{sec:strurepft}
\strurepft{} {\cite{LoRAMoE_ACL24,MoSLoRA_EMNLP24,MixLoRA_ACL24,FourierFT_ICML24,LoRArar_arXiv24,TeamLoRA_arXiv24,REMEDY_ICLR25}} focuses on reorganizing or reconfiguring model architectures through additional architecture or dynamic experts adjustments to boost \lora{} efficiency.

\noindent \textbf{Additional Architecture Design.} A substantial body of work focuses on enhancing model architecture to uncover knowledge patterns. For example, \vera{} \cite{VERA_ICLR24} freezes a single pair of randomly initialized matrices and introduces trainable scaling vectors. \moslora{} \cite{MoSLoRA_EMNLP24} employs a learnable mixer to fuse multiple subspaces. \mtlora{} \cite{MTLoRA_CVPR24} proposes both task-agnostic and task-specific low-rank adaptations. \fourierft{} \cite{FourierFT_ICML24} treats $\Delta W$ as a matrix in the spatial domain and learns only a small fraction of its spectral coefficients. \mixlora{} \cite{MixLoRA_ACL24} employs a dynamic factor selection mechanism, which includes independent and conditional factor selection routers tailored to the unique demands of each input instance. This line of research focuses on disassembling \lora{} into various specialized variants.

\noindent \textbf{Dynamical Expert Combination.} Another line of research explores multidimensional task scenarios \cite{TeamLoRA_arXiv24,LoRAMoE_ACL24,ShareLoRA_arXiv24,REMEDY_ICLR25,TwinMerging_NeurIPS24} and aims to integrate knowledge from multiple parties. A straightforward approach involves applying the Mixture of Experts (MoE) theory \cite{AdaptiveMoE_NeurComput91,MoEforMulitClass_NeurNet99, SparseMoE_arXiv17,MoE_AIR14,MoEMeetsLLMS_SIGIR24}, which adaptively integrates the diverse knowledge of multiple \lora{} experts to address the varying characteristics of different tasks. Early works typically treat each \lora{} module as an individual expert, as seen in \loramoe{} \cite{LoRAMoE_ACL24}. Furthermore, \teamlora{} \cite{TeamLoRA_arXiv24} and \sharelora{} \cite{ShareLoRA_arXiv24} explore asymmetric collaboration. Specifically, these methods share the A matrix to capture homogeneous features for general knowledge, while learning distinct B matrices that focus on task-specific features. Recently, both \twinmerge{} \cite{TwinMerging_NeurIPS24} and \remedy{} \cite{REMEDY_ICLR25} have optimized the expert selection router for tuning \lora{} modules. In particular, \twinmerge{} modularize knowledge into shared and exclusive components through singular value decomposition. These approaches primarily address data conflicts and dynamically combine multi-party knowledge to adapt to varying distributions.

\subsubsection{\calirepft}
\label{sec:calirepft}
\calirepft{} {\cite{DOR_arXiv24,LoRAvsFull_arXiv24,CLoRA_arXiv24,Flora_ICML24,LaplaceLoRA_ICLR24}} turns to analyze  potentially inherent mechanisms for \lora{} and consider to introduce regularization term or modify the optimization paradigm as follows.

\noindent \textbf{Parameter Update Regularization}. This direction consider additional regularization term besides the normal fitting objective to prevent catastrophic forgetting and boost downstream performance. \clora{} \cite{CLoRA_arXiv24} introduces the  subspace regularization method on \lora structure. \milora{} \cite{MiLoRA_arXiv24}  learns on tuning tasks while  preserving the pretrained knowledge by adapting the minor singular components of pretrained weight matrices. \pego{} \cite{PEGO_ECCV24} apply an orthogonal regularization loss between the pre-trained weights and ongoing \lora{} module. \laplacelora{} \cite{LaplaceLoRA_ICLR24} utilizes Bayesian approach {\cite{WeightUncertaintyBayesian_ICML15,MCMCBayesian_arXiv19,BayesianRelu_ICML20,GlobalBayesian_ICML21}} to estimate uncertainty, serve as potent tools to mitigate overconfidence and enhance calibration. \pissa{} {\cite{PiSSA_NeurIPS24}}  updates the principal components while freezing the “residual” parts. \corda{} \cite{CorDA_NeurIPS24} performs SVD on pre-trained weights, guided by the covariance matrix, to capture task-specific information and aggregates the context into the principal components for maintenance or adaptation. Recently, \lorascruplt{} {\cite{LoRASculpt_CVPR25}} introduces sparse updates into LoRA and integrates pre-trained weights for knowledge-preserving regularization, enhancing general and downstream task knowledge fusion.

\noindent \textbf{Adaptive Optimization Strategies.} These techniques redesign optimization dynamics for improving  convergence and accuracy performance. \loraplus{} \cite{LoraPLUS_ICML24}  employs differentiated learning rates between projection matrices. Specifically, set the learning rates for $A$,$B$ such that $\eta_B = \lambda \eta_A$ with $\lambda>1$. \bilora{} \cite{BiLoRA_arXiv24} implements bi-level optimization and separately trains pseudo singular vectors on distinct sub-datasets in two different optimization levels.  We offer a overview for current solution in \cref{tab:repft_summary}.

\begin{table}[t]\small
\caption{\textbf{Summary of essential characteristics for existing works in \repft{} (\cref{sec:repft})}. We measure whether the proposed method could be merged back to original architecture.}
\label{tab:repft_summary}
\vspace{-10pt}
\centering
{
\resizebox{\columnwidth}{!}{
\setlength\tabcolsep{2pt}
\renewcommand\arraystretch{1.2}
\begin{tabular}{rIc|c|c}
\hline\thickhline
\rowcolor{mydarkgreen}
\textbf{Methods} & \textbf{Venue} & \textbf{Highlight} & \textbf{Merge}  \\
\hline\hline
\multicolumn{4}{l}{\textit{\textbf{Structure \repft{}}}} \\  
\hline
\rowcolor{mylightgreen}
\loramoe{} \cite{LoRAMoE_ACL24} & \pub{ACL'24} &  Multiple LoRAs as adaptable experts  &\\
\moslora{} \cite{MoSLoRA_EMNLP24}  & \pub{EMNLP'24} & A learnable mixer for fusion &{\color{mydarkdarkgreen}\faCheck} \\ 
\rowcolor{mylightgreen}
\mixlora{} \cite{MixLoRA_ACL24}   & \pub{ACL'24} & Dynamic factor selection &   \\ 
\fourierft{} \cite{FourierFT_ICML24}  & \pub{ICML'24} & Weight changes as spatial-domain matrices &{\color{mydarkdarkgreen}\faCheck}\\
\rowcolor{mylightgreen}
\lorarar{} \cite{LoRArar_arXiv24}  & \pub{arXiv'24} & Pre-trained Hyper-network &{\color{mydarkdarkgreen}\faCheck}  \\ 
\teamlora{} \cite{TeamLoRA_arXiv24}  & \pub{arXiv'24} & Asymmetric collaboration and competition & \\
\rowcolor{mylightgreen}
\lorahub{} \cite{LoraHub_COLM24} & \pub{COLM'24} & Shiwa evolver searched &{\color{mydarkdarkgreen}\faCheck}  \\ 
\vera{} \cite{VERA_ICLR24} & \pub{ICLR'24} & Vector based random matrix adaptation & {\color{mydarkdarkgreen}\faCheck}\\
\rowcolor{mylightgreen}
\mtlora{} \cite{MTLoRA_CVPR24} & \pub{CVPR'24} & Task-agnostic and -specific LoRA &  \\
\sharelora{} \cite{ShareLoRA_arXiv24} & \pub{arXiv'24} & Shared low rank adaptation & {\color{mydarkdarkgreen}\faCheck} \\
\rowcolor{mylightgreen}
\twinmerge{} \cite{TwinMerging_NeurIPS24} &  \pub{NeurIPS'24} & Dynamically merge specialized knowledge &  \\
\remedy{} \cite{REMEDY_ICLR25} & \pub{ICLR'25} & Modality-aware expert allocator & \\
\hline\hline
\multicolumn{4}{l}{\textit{\textbf{Calibration \repft{}}}} \\  
\hline
\rowcolor{mylightgreen}
\clora{} \cite{CLoRA_arXiv24} & \pub{arXiv'24} & Subspace regularization &{\color{mydarkdarkgreen}\faCheck}  \\
\hiddenkey{} \cite{HiddenKey_ACL24}  & \pub{ACL'24} & Drop columns and elements of attention &{\color{mydarkdarkgreen}\faCheck}  \\
\rowcolor{mylightgreen}
\loraplus{} \cite{LoraPLUS_ICML24} & \pub{ICML'24} & Different learning rates  &{\color{mydarkdarkgreen}\faCheck}  \\
\dora{} \cite{DoRA_ICML24} & \pub{ICML'24} & Weight decomposed low-rank  &{\color{mydarkdarkgreen}\faCheck}  \\
\rowcolor{mylightgreen}
\flora{} \cite{Flora_ICML24}& \pub{ICML'24} & Resample projection matrices  &{\color{mydarkdarkgreen}\faCheck}  \\
\corda{} \cite{CorDA_NeurIPS24}  & \pub{NeurIPS'24} & Preserve knowledge, preview instruction  & {\color{mydarkdarkgreen}\faCheck}  \\
\rowcolor{mylightgreen}
\milora{}  \cite{MiLoRA_arXiv24} & \pub{arXiv'24} & Adapting  minor singular components & {\color{mydarkdarkgreen}\faCheck}  \\
\pissa{} \cite{PiSSA_NeurIPS24} & \pub{NeurIPS'24} &  Principal and residual components & {\color{mydarkdarkgreen}\faCheck}  \\
\rowcolor{mylightgreen}
\melora{} \cite{MELoRA_ACL24} & \pub{ACL'24} & Stack multiple mini LoRAs  & {\color{mydarkdarkgreen}\faCheck}  \\
\prolora{} \cite{PRoLoRA_ACL24} & \pub{ACL'24} & Intra-layer sharing via partial rotation &{\color{mydarkdarkgreen}\faCheck}  \\
\rowcolor{mylightgreen}
\pego{} \cite{PEGO_ECCV24} & \pub{ECCV'24} & Orthogonal group regularization &{\color{mydarkdarkgreen}\faCheck}  \\
\bilora{} \cite{BiLoRA_arXiv24}  &  \pub{arXiv'24} & Bi-level optimize via pseudo SVD \\
\rowcolor{mylightgreen}
\laplacelora{} \cite{LaplaceLoRA_ICLR24} & \pub{ICLR'24} & Laplace approx to LoRA posterior  & \\
\lorascruplt{} \cite{LoRASculpt_CVPR25} & \pub{CVPR'25} & Sparse updates with knowledge protection & {\color{mydarkdarkgreen}\faCheck}  \\
\hline\thickhline
\end{tabular}}}
\vspace{-15pt}
\end{table}

\subsubsection{Pros and Cons}
We discuss the advantages for the \repft{} as follows.
\begin{itemize}
\item[\faThumbsOUp] \textbf{Adapting Heterogeneous Architecture.} \repft{} operates on the standard matrices and decomposes them into low-rank properties, offering high flexibility across different model architectures.

\item[\faThumbsOUp] \textbf{Saving Computational Resources.} \lora{} saves memory and computational resources by training only low-rank perturbations of selected weight matrices. By freezing the majority of the original model weights and only updating low-rank matrices, \lora{} minimizes the memory footprint required during training. This makes  \lora{} particularly well-suited for resource-constrained environments or scenarios with limited computational power.
\end{itemize}
We further list the relevant drawbacks for the \repft{} methods.
\begin{itemize}
\item[\faThumbsODown] \textbf{Representation Ability Limitation.} The low-rank assumption restricts the expressiveness ability \cite{LoRAvsFull_arXiv24,LoraLearnForget_TMLR24}. This limitation becomes particularly apparent when the model needs to capture intricate patterns or high-dimensional relationships that cannot be adequately represented by low-rank matrices.

\item[\faThumbsODown] \textbf{Sensitivity to Rank Behavior.} \lora{} involves a large number of hyper-parameters to select, including target modules, rank, scaling factors, and learning rates. Specifically, the rank scale serves as a crucial hyper-parameter for \lora{}  expressive capacity \cite{LoraLearnForget_TMLR24}.
\end{itemize}

\section{Benchmark}
\label{sec:banchmark}

\subsection{Setup}
\label{sec:setup}
\subsubsection{Datasets}
We categorize the datasets into two groups: pre-training (seen) and downstream-tuning (unseen) datasets, in order to assess both the generalization and specialization abilities. The pre-training datasets consist of those used during the training process. To evaluate the learned generalization ability, we use the following datasets: \okvqa{} \cite{OKVQA_CVPR19}, \gqa{} \cite{GQA_CVPR19}, \textvqa{} \cite{TextVQA_CVPR19}, \ocrvqa{} \cite{OCRVQA_ICDAR19}, \cococap{} \cite{COCO_ECCV14}, and \mme{} \cite{MME_ICLR25}.
The first five VQA and captioning datasets are used to assess source-domain capabilities of MLLMs, while MME is employed to evaluate the retention of diverse world knowledge.
For downstream tasks, we consider tasks from diverse domains, which include: \scienceqa{} \cite{ScienceQA_NeurIPS22}, \iconqa{} \cite{IconQA_NeurIPS21}, \refcoco{} \cite{RefCOCO_EMNLP14}, \icfg{} \cite{ICFG_arXiv21}, and \rsvqa{} \cite{RSVQA_TGRS20}. A detailed introduction to these datasets is provided in \cref{tab:dataset_introduction}. For tasks involving caption responses, we use CIDEr as the evaluation metric, while for other tasks, we employ the standard classification metric.

\subsubsection{Architecture and Tuning Methods}
Adhering to the \mllm{} paradigm, we evaluate the effectiveness of our methods using two popular models as the foundation for our experiments: \llavaov{} \cite{LLaVAOneVision_arXiv24,LLaVA_NeurIPS23} and \vila{} \cite{VILA_CVPR24}. We utilize \llavaov{}-QWEN2-7b-Si \cite{QwenVL_arXiv23,LLaVANext_arXiv24} and \vila{}-3B in our experiments. In our work, we focus on the {\color{mydarkdarkred}\selft{}} and {\color{mydarkdarkgreen}\repft{}} paradigms, which have been empirically confirmed as effective tuning methods in the \mllmabbrv{} era with  high architecture transferability {\cite{LoRAvsFull_arXiv24}}. However, \addft{} interrupts with the model architecture and bring less transferability across tasks. As  for \selft{} paradigm, we consider the computation resource restriction setting, and further solely tune the \textit{Top} and \textit{Last}  $L$ blocks layers for experiments. We further illustrate the candidate tuning methods:
\begin{fullitemize}
\item \lora{} \cite{LoRA_ICLR22}: Applies low-rank matrix adaptation for parameter-efficient tuning.

\item \dora{} \cite{DOR_arXiv24}: Decomposing weights into magnitude and direction for improved parameter-efficient tuning.

\item \fullselft{} (\fullselftabbrv): Tuning all layers without additional structural constraints.

\item  \topselft{} (\topselftabbrv): Tuning only the uppermost layers of \llmabbrv{} while others frozen.

\item  \lastselft{} (\lastselftabbrv): Tuning only the final layers of \llmabbrv{} while others frozen.
\end{fullitemize}

\begin{table*}[t]\small
\caption{\textbf{Detailed Dataset Description}.}
\label{tab:dataset_introduction}
\vspace{-10pt}
\centering
{
\resizebox{\textwidth}{!}{
\setlength\tabcolsep{2pt}
\renewcommand\arraystretch{1.2}
\begin{tabular}{rIc|c|c|c|c|cc}
\hline\thickhline
\rowcolor{mydarkyellow}
Dataset & Venue & Task & Metric  & Answer &  Prompt & \multicolumn{2}{|c}{Description} \\
\hline\hline
\multicolumn{4}{l}{\textit{Upstream Datasets} $\mathcal{S}$} \\  
\hline
\cococap{} \cite{COCO_ECCV14} & \pub{ECCV'14} & Common Objects Caption & CIDER ($\uparrow$)  & Caption & \multicolumn{1}{c|}{\makecell[c]{Provide a one-sentence caption\\ for the provided image.}} & \protect\includegraphics[scale=0.06,valign=c]{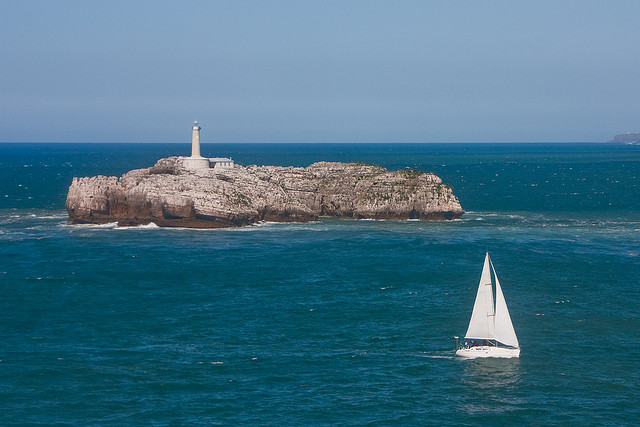} 
 & \multicolumn{1}{c}{\makecell[c]{\underline{\textbf{A}}: A sailboat is sailing \\ in the ocean.}} \\ 

\okvqa{} \cite{OKVQA_CVPR19} & \pub{CVPR'19} & Outside-knowledge VQA & Accuracy ($\uparrow$) & Phrase  & \multicolumn{1}{c|}{\makecell[c]{Answer the question using \\ a single word or phrase.}}
& \protect\includegraphics[scale=0.06,valign=c]{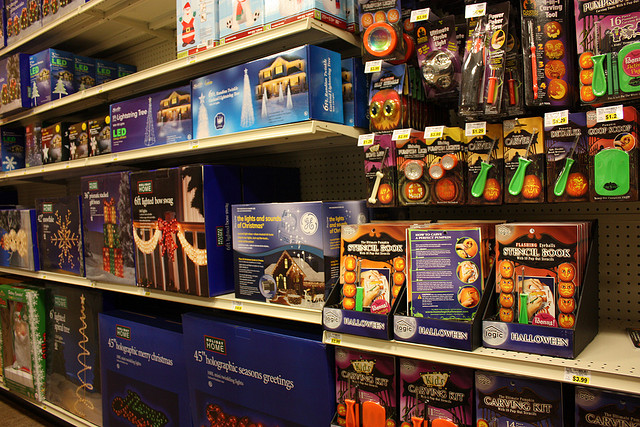} 
 & \multicolumn{1}{c}{\makecell[c]{\underline{\textbf{Q}}: What place is this? \\ \underline{\textbf{A}}: Store.}}\\
 
\textvqa{} \cite{TextVQA_CVPR19} & \pub{CVPR'19} & Reading Comprehension VQA & Accuracy ($\uparrow$) & Phrase  & \multicolumn{1}{c|}{\makecell[c]{Answer the question using \\ a single word or phrase.}} & \protect\includegraphics[scale=0.04,valign=c]{Figure/TextVQA_Example.jpg} 
 &\multicolumn{1}{c}{\makecell[c]{\underline{\textbf{Q}}: Is this denny's? \\ \underline{\textbf{A}}: Yes.}}\\
 
\gqa{} \cite{GQA_CVPR19} & \pub{CVPR'19} & Image Scene Graphs VQA   & Accuracy ($\uparrow$)  & Phrase & \multicolumn{1}{c|}{\makecell[c]{Answer the question using \\ a single word or phrase.}} & \protect\includegraphics[scale=0.25,valign=c]{Figure/GQA_Example.jpg} 
 &\multicolumn{1}{c}{\makecell[c]{\underline{\textbf{Q}}: Is the snow bright? \\ \underline{\textbf{A}}: Yes.}} \\

\ocrvqa{} \cite{OCRVQA_ICDAR19} & \pub{CVPR'19} & VQA by Reading Text  & Accuracy ($\uparrow$)  & Phrase & \multicolumn{1}{c|}{\makecell[c]{Answer the question using \\ a single word or phrase.}}  & \protect\includegraphics[scale=0.05,valign=c]{Figure/OCRVQA_Example.jpg} 
 &\multicolumn{1}{c}{\makecell[c]{\underline{\textbf{Q}}:  Who wrote this book? \\ \underline{\textbf{A}}: Simon Hill.}} \\

\mme{} \cite{MME_arXiv23} & \pub{arXiv'23} & \multicolumn{1}{c|}{\makecell[c]{Real-world applications \\  with practical relevance}}   & Accuracy ($\uparrow$) & Phrase  &\multicolumn{1}{c|}{\makecell[c]{Answer the question using \\ a single word or phrase.}} & \protect\includegraphics[scale=0.05,valign=c]{Figure/MME_Example.jpg} 
 &\multicolumn{1}{c}{\makecell[c]{\underline{\textbf{Q}}: Is this an image of Guozijian? \\ \underline{\textbf{A}}: Yes.}}   \\

\hline\hline
\multicolumn{4}{l}{\textit{Downstream Datasets} $\mathcal{T}$ (Train/Test)}  \\  
\hline
\multicolumn{1}{rI}{\makecell[r]{\flickrthirtyk{}  \cite{Flickr_TACL14} \\ $(10000/1000)$}} & \pub{TACL'14} & Everyday activities portrayal & CIDER ($\uparrow$) & Caption & \multicolumn{1}{c|}{\makecell[c]{Provide a one-sentence caption\\ for the provided image.}}  & \protect\includegraphics[scale=0.12,valign=c]{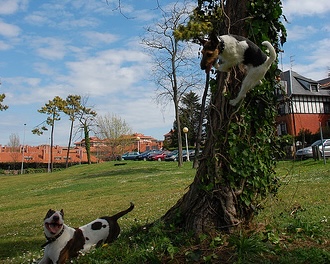} 
 &\multicolumn{1}{c}{\makecell[c]{\underline{\textbf{A}}: A dog jumps by a tree while \\ another lays on the ground.}}  
 \\
 
\multicolumn{1}{rI}{\makecell[r]{\rsvqa{} \cite{RSVQA_TGRS20}  \\ $(10000/10004)$}}   & \pub{TGRS'20} & VQA for Remote Sensing & Accuracy ($\uparrow$) & Phrase & \multicolumn{1}{c|}{\makecell[c]{Answer the question using \\ a single word or phrase.}} & \protect\includegraphics[scale=0.15,valign=c]{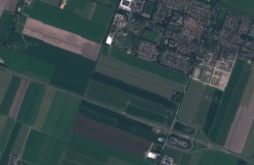}
 &\multicolumn{1}{c}{\makecell[c]{Q: Is there a road? \\ \underline{\textbf{A}}: Yes.}}  
 \\

\multicolumn{1}{rI}{\makecell[r]{\pathvqa{} \cite{PathVQA_arXiv20}  \\ $(10000/6719)$}}   & \pub{arXiv'20} &  Pathology images  VQA & Accuracy ($\uparrow$) & Phrase & \multicolumn{1}{c|}{\makecell[c]{Answer the question using \\ a single word or phrase.}}  & \protect\includegraphics[scale=0.05,valign=c]{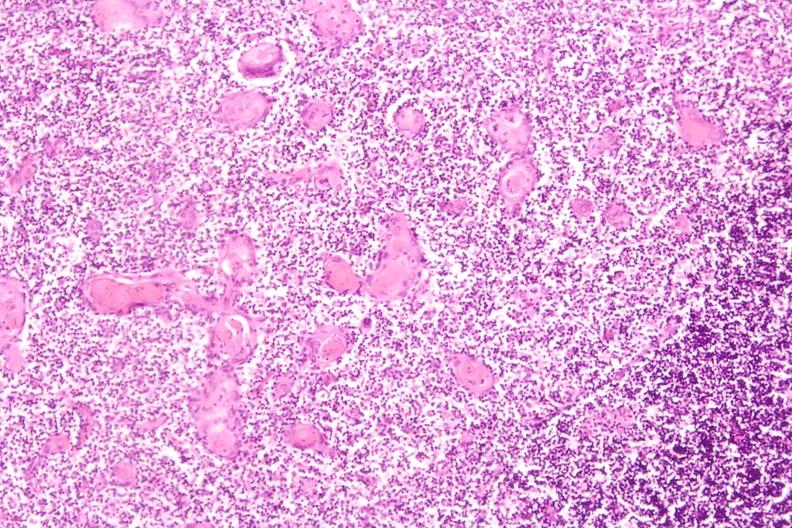}
 &\multicolumn{1}{c}{\makecell[c]{\underline{\textbf{Q}}: Does this image show thymus? \\ \underline{\textbf{A}}: Yes}}    \\

\multicolumn{1}{rI}{\makecell[r]{\iconqa{} \cite{IconQA_NeurIPS21}  \\ $(10000/6316)$}}   & \pub{NeurIPS'21} & Abstract Diagram Understanding & Accuracy ($\uparrow$) & Option & \multicolumn{1}{c|}{\makecell[c]{Answer with the option's letter \\ from the given choices directly.}}  & \protect\includegraphics[scale=0.05,valign=c]{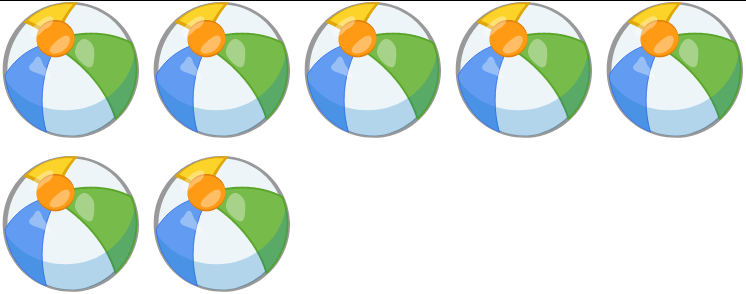}
 &\multicolumn{1}{c}{\makecell[c]{\underline{\textbf{Q}}: How many balls are there? \\ A. 1 B. 3 C. 8 D. 7 E. 2  \\ \underline{\textbf{A}}: D}}  \\

\multicolumn{1}{rI}{\makecell[r]{\scienceqa{} \cite{ScienceQA_NeurIPS22} \\ $(6218/2017)$}}   & \pub{NeurIPS'22}  & Science Question Answering  & Accuracy ($\uparrow$) & Option & \multicolumn{1}{c|}{\makecell[c]{Answer with the option's letter \\ from the given choices directly.}} & \protect\includegraphics[scale=0.05,valign=c]{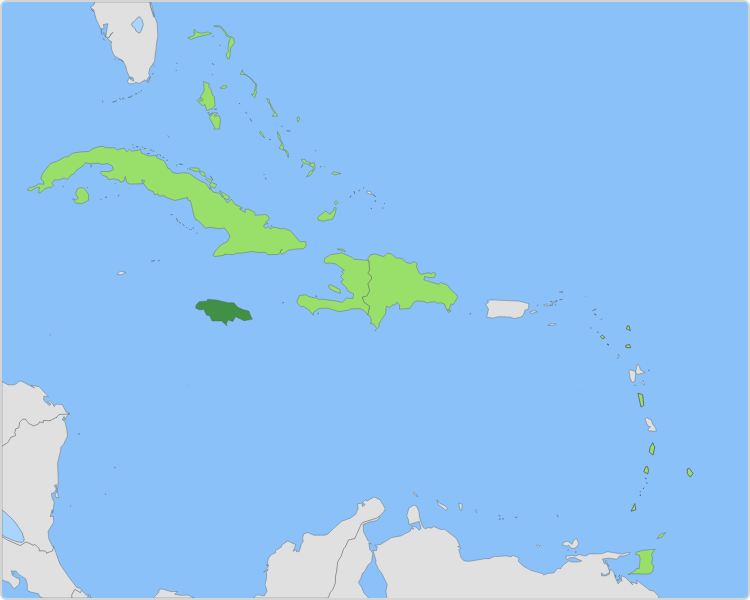}
 &\multicolumn{1}{c}{\makecell[c]{\underline{\textbf{Q}}: Which country is highlighted? \\ A. Saint Lucia B. Jamaica C. Haiti D. Cuba  \\ \underline{\textbf{A}}: D}}  \\

\multicolumn{1}{rI}{\makecell[r]{CHD \\ $(5373/2000)$}} & \pub{Private'25} & \multicolumn{1}{c|}{\makecell[c]{Early pregnancy fetal \\ cardiac ultrasound VQA}}   & F1 ($\uparrow$) & Option & \multicolumn{1}{c|}{\makecell[c]{Which of the following descriptions most  \\ accurately reflects the diagnostic result \\ of the image?  Answer with the option's \\ letter  from the given choices directly.}} & \protect\includegraphics[scale=0.15,valign=c]{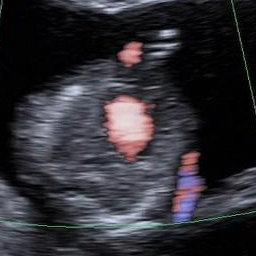}
 &\multicolumn{1}{c}{\makecell[c]{\underline{\textbf{Q}}: A. Single ventricle B. Atrioventricular \\ septal defect  C. Atrioventricular  valve  \\ atresia D. Ventricular hypoplasia  \\ E. Tricuspid valve dysplasia F. Normal \\ \underline{\textbf{A}}: D}} 
\end{tabular}}}
\end{table*}

\definecolor{lightpink}{RGB}{253, 244, 241}

\definecolor{lightblue}{RGB}{245, 255, 255}

\definecolor{lightgreen}{RGB}{255, 255, 241}

\begin{table*}[t]\small
\captionsetup{font=small}
\caption{{
\textbf{Comparison with Representative \mllm{} (\mllmabbrv{}) Tuning Solutions} on different downstream datasets $\mathcal{T}$ based on  the \textbf{\llavaov{}} architecture. We mark the Best in bold across different tuning methods.  {\color{myorange}$\uparrow$} and {\color{mygreen}$\downarrow$} means improvement and decrease ratio compared with ZS (Zero-shot).  Please refer to \cref{sec:exp_com} for relative explanations. 
}}
\label{tab:compare_sota_llava}
\vspace{-10pt}
\centering
\scriptsize{
\resizebox{\linewidth}{!}{
\setlength\tabcolsep{1.pt}
\renewcommand\arraystretch{1.2}
\begin{tabular}{>{\columncolor{mylightgray}}r||>{\columncolor{mylightgray}}c>{\columncolor{mylightgray}}c>{\columncolor{mylightgray}}c|>{\columncolor{lightpink}}c>{\columncolor{lightblue}}c>{\columncolor{lightgreen}}cI>{\columncolor{mylightgray}}c>{\columncolor{mylightgray}}c>{\columncolor{mylightgray}}c|>{\columncolor{lightpink}}c>{\columncolor{lightblue}}c>{\columncolor{lightgreen}}cI>{\columncolor{mylightgray}}c>{\columncolor{mylightgray}}c>{\columncolor{mylightgray}}c|>{\columncolor{lightpink}}c>{\columncolor{lightblue}}c>{\columncolor{lightgreen}}c}
\hline\thickhline
\rowcolor{gray!20}
 & 
\multicolumn{6}{cI}{\bm{\scienceqa{}}} & 
\multicolumn{6}{cI}{\bm{{\iconqa{}}}} &
\multicolumn{6}{c}{\bm{{CHD}}}
\\
\cline{2-19} 
\rowcolor{gray!20}
 
& VQA & Cap & \mme{} & $\mathcal{F}$ & $\mathcal{A}^\mathcal{T}$  & $\mathcal{O}$
& VQA & Cap & \mme{} & $\mathcal{F}$ & $\mathcal{A}^\mathcal{T}$  & $\mathcal{O}$
& VQA & Cap & \mme{} & $\mathcal{F}$ & $\mathcal{A}^\mathcal{T}$  & $\mathcal{O}$
\\
\hline\hline

ZS 
& 66.71 &	140.14	&	1,693.87/513.93	&	- & 78.24 & -	
& 66.71 &	140.14	&	1,693.87/513.93	&	- & 43.33 & -	
& 66.71 &	140.14	&	1,693.87/513.93	&	- & 3.32 & -	

\\
\hdashline
\multicolumn{16}{l}{\textbf{\textit{\repft{}}}} 
\\

\lora{} 
& \tworowup{67.32}{0.92} & \tworowdown{140.00}{-0.10} & \tworowdown{1,710.00/492.50}{-1.61} & -0.26 & \tworowup{88.55}{13.18} & 12.92
& \tworowup{67.57}{1.30} & \tworowdown{140.13}{-0.01} & \tworowdown{1,669.00/496.07}{-2.47} & \textbf{-0.39} & \tworowup{85.36}{120.08} & 119.68
& \tworowdown{66.65}{-0.09} & \tworowup{140.26}{0.09} & \tworowdown{1,695.85/509.29}{-0.39} & \textbf{-0.13} & \tworowup{77.58}{2,236.75} & 2,236.61\\

\dora{} 
& \tworowup{67.41}{1.05} & \tworowdown{139.99}{-0.11} & \tworowdown{1,706.25/493.21}{-1.65} & \textbf{-0.23} & \tworowup{90.98}{16.28} & \textbf{16.05}
& \tworowup{67.55}{1.26} & \tworowup{140.22}{0.06} & \tworowdown{1,670.35/481.07}{-3.89} & -0.86 & \tworowup{95.42}{120.22} & 119.36
& \tworowdown{66.64}{-0.10} & \tworowup{140.17}{0.02} & \tworowdown{1,694.96/501.79}{-1.15} & -0.41 & \tworowup{78.00}{2,249.40} & 2,248.99\\

\hdashline
\multicolumn{16}{l}{\textbf{\textit{\selft}}} 
\\ 
Full
& \tworowdown{54.26}{-18.66} & \tworowdown{133.29}{-4.89} & \tworowdown{942.98/125.00}{-60.00} & -27.85 & \tworowup{\textbf{96.93}}{23.89} & -3.96
& \tworowdown{24.31}{-63.56} & \tworowdown{125.74}{-10.28} & \tworowdown{50.92/3.57}{-98.15} & -57.33 & \tworowup{\textbf{97.51}}{125.04} & 67.71
& \tworowdown{36.36}{-45.50} & \tworowdown{127.09}{-9.31} & \tworowdown{623.90/30.36}{-78.63} & -44.48 & \tworowup{\textbf{81.32}}{2,349.40} & \textbf{2,304.92}\\

Top
& \tworowdown{65.20}{-2.27} & \tworowdown{139.71}{-0.31} & \tworowdown{1,722.93/486.07}{-1.85} & -1.48 & \tworowup{91.47}{16.91} & 15.43
& \tworowdown{65.34}{-2.05} & \tworowdown{139.30}{-0.60} & \tworowdown{1,704.56/473.21}{-3.65} & -2.10 & \tworowup{96.93}{123.70} & \textbf{121.60}
& \tworowdown{64.25}{-3.69} & \tworowdown{139.15}{-0.71} & \tworowdown{1,685.51/489.29}{-2.64} & -2.35 & \tworowup{77.58}{2,236.75} & 2,234.40\\

Last
& \tworowdown{64.39}{-3.48} & \tworowdown{138.76}{-0.98} & \tworowdown{1,716.06/480.71}{-2.58} & -2.35 & \tworowup{85.42}{9.18} & 6.83
& \tworowdown{58.97}{-11.60} & \tworowdown{137.38}{-1.97} & \tworowdown{1,676.79/500.71}{-1.79} & -5.12 & \tworowup{93.43}{115.62} & 110.5
& \tworowdown{57.85}{-13.29} & \tworowdown{137.76}{-1.70} & \tworowdown{1,690.40/501.43}{-1.32} & -5.43 & \tworowup{76.32}{2,198.80} & 2,193.36\\

\hline\thickhline
\rowcolor{gray!20}
 & 
\multicolumn{6}{cI}{\bm{\pathvqa{}}} & 
\multicolumn{6}{cI}{\bm{{\rsvqa{}}}} &
\multicolumn{6}{c}{\bm{{\flickrthirtyk{}}}}
\\
\cline{2-19} 
\rowcolor{gray!20}
 
& VQA & Cap & \mme{} & $\mathcal{F}$ & $\mathcal{A}^\mathcal{T}$  & $\mathcal{O}$
& VQA & Cap & \mme{} & $\mathcal{F}$ & $\mathcal{A}^\mathcal{T}$  & $\mathcal{O}$
& VQA & Cap & \mme{} & $\mathcal{F}$ & $\mathcal{A}^\mathcal{T}$  & $\mathcal{O}$
\\
\hline\hline

ZS 
& 66.71 & 140.14 & 1,693.87/513.93 & - & 45.78 & -
& 66.71 & 140.14 & 1,693.87/513.93 & - & 52.56 & - 
& 66.71 & 140.14 & 1,693.87/513.93 & - & 81.59 & -

\\
\hdashline
\multicolumn{16}{l}{\textbf{\textit{\repft{}}}} 
\\ 

\lora{} 
& \tworowdown{65.69}{-1.53} & \tworowdown{138.59}{-1.11} & \tworowdown{1,687.84/465.71}{-4.87} & \textbf{-2.50} & \tworowup{59.90}{30.84} & 28.34 
& \tworowdown{64.03}{-4.01} & \tworowdown{139.33}{-0.58} & \tworowdown{1,682.30/473.93}{-4.23} & -2.94 & \tworowup{72.03}{37.04} & 34.10 
& \tworowdown{66.54}{-0.25} & \tworowdown{114.98}{-17.95} & \tworowdown{1,691.46/472.86}{-4.07} & \textbf{-7.43} & \tworowup{92.37}{13.21} & 5.79 \\

\dora{} 
& \tworowdown{65.59}{-1.68} & \tworowdown{138.41}{-1.23} & \tworowdown{1,697.53/460.36}{-5.10} & -2.67 & \tworowup{60.10}{31.28} & \textbf{28.61} 
& \tworowdown{64.03}{-4.02} & \tworowdown{139.41}{-0.52} & \tworowdown{1,687.44/478.57}{-3.63} & \textbf{-2.72} & \tworowup{72.13}{37.23} & \textbf{34.51}
& \tworowdown{66.64}{-0.10} & \tworowdown{115.34}{-17.70} & \tworowdown{1,681.73/471.07}{-4.53} & -7.44 & \tworowup{92.61}{13.51} & 6.07 \\

\hdashline
\multicolumn{16}{l}{\textbf{\textit{\selft}}} 
\\ 
Full
& \tworowdown{54.45}{-18.38} & \tworowdown{105.40}{-24.79} & \tworowdown{1,479.71/336.07}{-23.63} & -22.27 & \tworowup{\textbf{62.18}}{35.82} & 13.56
& \tworowdown{53.80}{-19.35} & \tworowdown{119.77}{-14.54} & \tworowdown{1,182.73/383.21}{-27.81} & -20.56 & \tworowup{\textbf{72.16}}{37.29} & 16.73
& \tworowdown{56.81}{-14.83} & \tworowdown{91.05}{-35.03} & \tworowdown{1,601.04/418.93}{-11.98} & -20.62 & \tworowdown{73.68}{-9.69} & -30.31\\

Top
& \tworowdown{58.47}{-12.35} & \tworowdown{133.00}{-5.09} & \tworowdown{1,632.16/440.71}{-8.95} & -8.80 & \tworowup{58.71}{28.24} & 19.45
& \tworowdown{59.25}{-11.18} & \tworowdown{136.81}{-2.38} & \tworowdown{1,600.87/423.21}{-11.57} & -8.38 & \tworowup{71.59}{36.21} & 27.83
& \tworowdown{62.80}{-5.86} & \tworowdown{118.12}{-15.71} & \tworowdown{1,667.24/472.86}{-4.78} & -8.79 & \tworowup{\textbf{93.94}}{15.14} & \textbf{6.35} \\

Last
& \tworowdown{62.87}{-5.76} & \tworowdown{129.76}{-7.41} & \tworowdown{1,477.88/298.93}{-27.29} & -13.49 & \tworowup{59.26}{29.45} & 15.96
& \tworowdown{63.09}{-5.43} & \tworowdown{138.11}{-1.45} & \tworowdown{1,575.79/388.93}{-15.65} & -7.51 & \tworowup{71.23}{35.52} & 28.01
& \tworowdown{63.39}{-4.97} & \tworowdown{105.44}{-24.76} & \tworowdown{1,666.61/471.07}{-4.97} & -11.57 & \tworowup{86.81}{6.40} & -5.17\\

\end{tabular}}}
\end{table*}

\begin{table*}[t]\small
\captionsetup{font=small}
\caption{{
\textbf{Comparison with Representative \mllm{} (\mllmabbrv{}) Tuning Solutions} on different downstream datasets $\mathcal{T}$ based on  the \textbf{\vila{}} architecture. We mark the Best in bold across different tuning methods.  {\color{myorange}$\uparrow$} and {\color{mygreen}$\downarrow$} means improvement and decrease ratio compared with ZS (Zero-shot).  Please refer to \cref{sec:exp_com} for relative explanations. 
}}
\label{tab:compare_sota_vila}
\vspace{-10pt}
\centering
\scriptsize{
\resizebox{\linewidth}{!}{
\setlength\tabcolsep{1.pt}
\renewcommand\arraystretch{1.2}
\begin{tabular}{>{\columncolor{mylightgray}}r||>{\columncolor{mylightgray}}c>{\columncolor{mylightgray}}c>{\columncolor{mylightgray}}c|>{\columncolor{lightpink}}c>{\columncolor{lightblue}}c>{\columncolor{lightgreen}}cI>{\columncolor{mylightgray}}c>{\columncolor{mylightgray}}c>{\columncolor{mylightgray}}c|>{\columncolor{lightpink}}c>{\columncolor{lightblue}}c>{\columncolor{lightgreen}}cI>{\columncolor{mylightgray}}c>{\columncolor{mylightgray}}c>{\columncolor{mylightgray}}c|>{\columncolor{lightpink}}c>{\columncolor{lightblue}}c>{\columncolor{lightgreen}}c}
\hline\thickhline
\rowcolor{gray!20}
 & 
\multicolumn{6}{cI}{\bm{\scienceqa{}}} & 
\multicolumn{6}{cI}{\bm{{\iconqa{}}}} &
\multicolumn{6}{c}{\bm{{CHD}}}
\\
\cline{2-19} 
\rowcolor{gray!20}
 
& VQA & Cap & \mme{} & $\mathcal{F}$ & $\mathcal{A}^\mathcal{T}$  & $\mathcal{O}$
& VQA & Cap & \mme{} & $\mathcal{F}$ & $\mathcal{A}^\mathcal{T}$  & $\mathcal{O}$
& VQA & Cap & \mme{} & $\mathcal{F}$ & $\mathcal{A}^\mathcal{T}$  & $\mathcal{O}$
\\
\hline\hline

ZS 
& 61.41	& 106.60 & 1,477.84/371.79 & - & 64.90 & - 
& 61.41	& 106.60 & 1,477.84/371.79 & - & 55.56 & - 
& 61.41	& 106.60 & 1,477.84/371.79 & - & 4.60 & - 

\\
\hdashline
\multicolumn{16}{l}{\textbf{\textit{\repft{}}}} 
\\

\lora{} 
& \tworowdown{60.65}{-1.25} & \tworowup{108.87}{2.13} & \tworowdown{1,401.42/262.86}{-17.23} & -5.45 & \tworowup{84.04}{29.49} & 24.04

& \tworowdown{53.49}{-12.90} & \tworowup{112.85}{5.86} & \tworowdown{1,266.15/173.57}{-33.82} & -13.62 & \tworowup{71.31}{28.35} & 14.73

& \tworowup{61.44}{0.04} & \tworowdown{103.79}{-2.64} & \tworowdown{1,411.02/386.43}{-0.29} & -0.96 & \tworowup{21.27}{362.39} & 361.43
\\

\dora{} 
& \tworowdown{60.17}{-2.03} & \tworowup{111.25}{4.36} & \tworowdown{1,433.14/274.64}{-14.58} & -4.08 & \tworowup{84.88}{30.79} & 26.70

& \tworowdown{52.14}{-15.10} & \tworowup{112.27}{5.32} & \tworowdown{1,020.71/167.86}{-42.89} & -17.56 & \tworowup{73.15}{31.66} & 14.10

& \tworowdown{60.77}{-1.05} & \tworowdown{101.66}{-4.63} & \tworowdown{1,379.21/356.79}{-5.35} & -3.68 & \tworowup{10.58}{130.00} & 126.32
\\
\hdashline
\multicolumn{16}{l}{\textbf{\textit{\selft}}} 
\\ 
Full
& \tworowdown{56.15}{-8.57} & \tworowup{107.25}{0.61} & \tworowdown{1,370.27/332.86}{-8.87} & -5.61 & \tworowup{\textbf{91.52}}{41.02} & 35.41

& \tworowdown{37.31}{-39.24} & \tworowup{110.23}{3.41} & \tworowdown{577.64/141.43}{-61.44} & -32.42 & \tworowup{\textbf{89.69}}{61.43} & 29.01

& \tworowdown{51.90}{-15.49} & \tworowdown{96.20}{-9.76} & \tworowdown{1,083.90/183.93}{-38.59} & -21.28 & \tworowup{\textbf{81.75}}{1,677.17} & \textbf{1,655.89}
\\
Top
& \tworowdown{61.12}{-0.48} & \tworowdown{104.57}{-1.90} & \tworowup{1,448.03/384.29}{0.67} & -0.57 & \tworowup{84.48}{30.17} & 29.60

& \tworowdown{60.70}{-1.16} & \tworowup{107.13}{0.50} & \tworowup{1,400.49/411.07}{2.67} & 0.67 & \tworowup{86.02}{54.82} & 55.49

& \tworowdown{60.78}{-1.03} & \tworowdown{106.15}{-0.42} & \tworowdown{1,448.09/362.86}{-2.21} & -1.22 & \tworowup{77.60}{1,586.96} & 1,585.74
\\

Last
& \tworowdown{60.49}{-1.50} & \tworowdown{102.33}{-4.01} & \tworowup{1,416.89/387.86}{0.10} & -1.80 & \tworowup{78.04}{20.25} & 18.44

& \tworowdown{60.16}{-2.04} & \tworowup{106.77}{0.16} & \tworowup{1,409.51/400.71}{1.58} & -0.10 & \tworowup{84.69}{52.43} & 52.33

& \tworowdown{61.07}{-0.55} & \tworowdown{105.26}{-1.26} & \tworowdown{1,465.39/372.14}{-0.37} & \textbf{-0.73} & \tworowup{75.40}{1,539.13} & 1,538.40
\\
\hline\thickhline
\rowcolor{gray!20}
 & 
\multicolumn{6}{cI}{\bm{{\pathvqa{}}}} & 
\multicolumn{6}{cI}{\bm{{\rsvqa{}}}} &
\multicolumn{6}{c}{\bm{{\flickrthirtyk{}}}}
\\
\cline{2-19} 
\rowcolor{gray!20}
 
& VQA & Cap & \mme{} & $\mathcal{F}$ & $\mathcal{A}^\mathcal{T}$  & $\mathcal{O}$
& VQA & Cap & \mme{} & $\mathcal{F}$ & $\mathcal{A}^\mathcal{T}$  & $\mathcal{O}$
& VQA & Cap & \mme{} & $\mathcal{F}$ & $\mathcal{A}^\mathcal{T}$  & $\mathcal{O}$
\\
\hline\hline

ZS 
& 61.41	& 106.60 & 1,477.84/371.79 & - & 29.13 & - 
& 61.41	& 106.60 & 1,477.84/371.79 & - & 52.77 & - 
& 61.41	& 106.60 & 1,477.84/371.79 & - & 76.51 & - 
\\
\hdashline
\multicolumn{16}{l}{\textbf{\textit{\repft{}}}} 
\\ 

\lora{} 
& \tworowdown{53.81}{-12.38} & \tworowdown{99.98}{-6.21} & 	\tworowdown{1,459.71/55.36}{-2.82} & -7.14 & \tworowup{55.19}{89.46} & 82.32

& \tworowdown{58.72}{-4.38} & \tworowup{108.92}{2.18}	&  \tworowdown{1,347.46/328.21}{-10.27} & -4.16 & \tworowup{70.28}{33.18} & 29.02

& \tworowdown{59.28}{-3.47} & \tworowdown{98.12}{-7.95} & \tworowdown{1,436.29/265.00}{-15.77} & -9.06 & \tworowup{80.07}{4.65} & -4.41
\\

\dora{} 
& \tworowdown{53.71}{-12.53} & \tworowdown{99.10}{-7.04} & \tworowdown{1,420.68/375.36}{-1.45} & -7.01 & \tworowup{55.13}{89.26} & 82.25

& \tworowdown{58.90}{-4.08} & \tworowup{108.63}{1.90} & \tworowdown{1,349.52/332.86}{-9.58} & 	-3.92 & \tworowup{70.22}{33.07} & 29.15

& \tworowdown{59.31}{-3.42} & \tworowdown{98.45}{-7.65} & \tworowdown{1,447.98/259.29}{-16.14}	& -9.07 & \tworowup{80.40}{5.08} & -3.98
\\

\hdashline
\multicolumn{16}{l}{\textbf{\textit{\selft}}} 
\\ 
Full
& \tworowdown{52.49}{-14.53} & \tworowup{108.46}{1.74} & \tworowdown{1,446.44/346.43}{-4.47} & 	-5.75 & \tworowup{\textbf{59.55}}{104.43} &	98.67

& \tworowdown{54.70}{-10.93} & 	\tworowdown{104.91}{-1.59} & \tworowdown{1,288.89/336.79}{-11.10} & -7.87 & \tworowup{\textbf{72.07}}{36.57} & 28.70

& \tworowdown{55.49}{-9.64} & \tworowdown{92.50}{-13.23} & \tworowdown{1,442.43/360.71}{-2.69}	& -8.52 & \tworowup{75.02}{-1.95} &	-10.46
\\

Top
& \tworowdown{57.58}{-6.23} & \tworowup{108.35}{1.64} & \tworowdown{1,410.22/356.07}{-4.40} & \textbf{-3.00} & \tworowup{56.51}{93.99} & \textbf{90.99}

& \tworowdown{59.68}{-2.82} & \tworowup{109.43}{2.65} & \tworowup{1,378.27/398.93}{0.28} & \textbf{0.04}
& \tworowup{71.09}{34.72} &	\textbf{34.76}

& \tworowdown{60.58}{-1.35} & \tworowup{113.98}{6.92} & \tworowdown{1,428.56/341.07}{-5.80} &	-0.07 & \tworowup{88.47}{15.63} & 15.56
\\

Last
&  \tworowdown{59.31}{-3.43} & \tworowdown{99.92}{-6.27} &	\tworowdown{1,424.42/301.07}{-11.32} & -7.00 &	\tworowup{54.68}{87.71} & 80.71

& \tworowdown{60.40}{-1.65} & \tworowdown{99.47}{-6.69} & \tworowup{1,413.90/411.43}{3.17} & 	-1.72 & \tworowup{69.65}{31.99} & 30.26

& \tworowdown{60.41}{-1.62} & \tworowup{111.33}{4.44} &	\tworowdown{1,457.81/362.14}{-1.98} &	\textbf{0.28} & \tworowup{\textbf{89.08}}{16.43} & \textbf{16.71}
\\

\end{tabular}}}
\end{table*}

\subsubsection{Evaluation Metrics}
To evaluate the performance of \mllm{} (\mllmabbrv{}) in both upstream generalization and downstream specialization aspects, we consider two key metrics: \textbf{Specialization Improvement} ($\mathcal{E}$) and \textbf{Stabilization Forgetting} ($\mathcal{F}$) Thus, we derive the following evaluation metrics forms:

\noindent\textbf{Specialization Improvement} ($\mathcal{E}$). We define the performance on the downstream task $\mathcal{T}$ and the performance of the learned $\theta$ and pre-trained $\theta^*$  as $\mathcal{A}_\mathcal{T} = Acc.(\theta, \mathcal{D}_\mathcal{T})$ and $\mathcal{A}^*_\mathcal{T} = Acc.(\theta^*, \mathcal{D}_\mathcal{T})$. Thus, the downstream specialization improvement after tuning is defined as:
\begin{equation}
\setlength\abovedisplayskip{2pt} \setlength\belowdisplayskip{2pt}
\mathcal{E} = \frac{\mathcal{A}_\mathcal{T} - \mathcal{A}_\mathcal{T}^*}{\mathcal{A}_\mathcal{T}^*}.
\label{eq:specialization}
\end{equation}
\noindent \textbf{Stabilization Forgetting} ($\mathcal{F}$). To quantify the generalization ability loss, Acc. denotes the accuracy metric. we define the pre-training source distribution as $\mathcal{S}=\{\mathcal{S}_i\}_{i=1}^{|\mathcal{S}|}$. We also denote the accuracy on optimized and default model as $\mathcal{A}_{\mathcal{S}_i} = Acc.(\theta, \mathcal{D}_{\mathcal{S}_i})$ and $\mathcal{A}^*_{\mathcal{S}_i} = Acc.(\theta^*, \mathcal{D}_{\mathcal{S}_i})$ Thus, the corresponding overall general knowledge forgetting $\mathcal{F}$ after \mllmabbrv{} tuning is defined as:
\begin{equation}
\setlength\abovedisplayskip{2pt} \setlength\belowdisplayskip{2pt}
\mathcal{F}_i  =  \frac{ \mathcal{A}_{\mathcal{S}_i}^* - \mathcal{A}_{\mathcal{S}_i}}{\mathcal{A}_{\mathcal{S}_i}^*}, \quad 
\mathcal{F}  = \frac{\sum_i^{|\mathcal{S}|} \mathcal{F}_i}{|\mathcal{S}|} .
\label{eq:forgetting}
\end{equation}
\noindent \textbf{Generalization and Stabilization Trade-Off.}
To comprehensively evaluate both the specialization ability and stabilization forgetting in \mllmabbrv{}, we use the O-Average metric ($\mathcal{O}$) \cite{ModelTailor_ICML24}. The O-Average metric measures the arithmetic mean of specialization improvement ($\mathcal{E}$) and generalization forgetting ($\mathcal{F}$) performance as follows:
\begin{equation}\small
\setlength\abovedisplayskip{2pt} \setlength\belowdisplayskip{2pt}
\mathcal{O} = \mathcal{E} + \mathcal{F}.
\label{eq:mean}
\end{equation}

\subsubsection{Implementation Details}
We follow the official codebase\footnote{https://github.com/LLaVA-VL/LLaVA-NeXT}$^{,}$\footnote{https://github.com/NVlabs/VILA} to conduct the tuning procedure. The default learning rate $lr$ is set to $1e-5$ for \llavaov{} {\cite{LLaVAOneVision_arXiv24}} and $1e-4$ for \vila{} {\cite{VILA_CVPR24}}. However, for \vila{} \fullselftabbrv{}, training was unstable with $lr = 1e-4$, so $lr$ is set to $1e-5$ to ensure stability.
The training epoch is set to $E=3$. The training batch size $B$ is set to 16. The maximum sequence length is set to 4096 for LLaVA-OV and VILA. As for \selft{}, the tuning block for \llmabbrv{} is the \textit{Top} and \textit{Last} $L=2$ layers. All experiments are conducted on 16 A100 GPUs, each with 80GB memory.

\begin{figure*}[h]
\centering
\includegraphics[width=\linewidth]{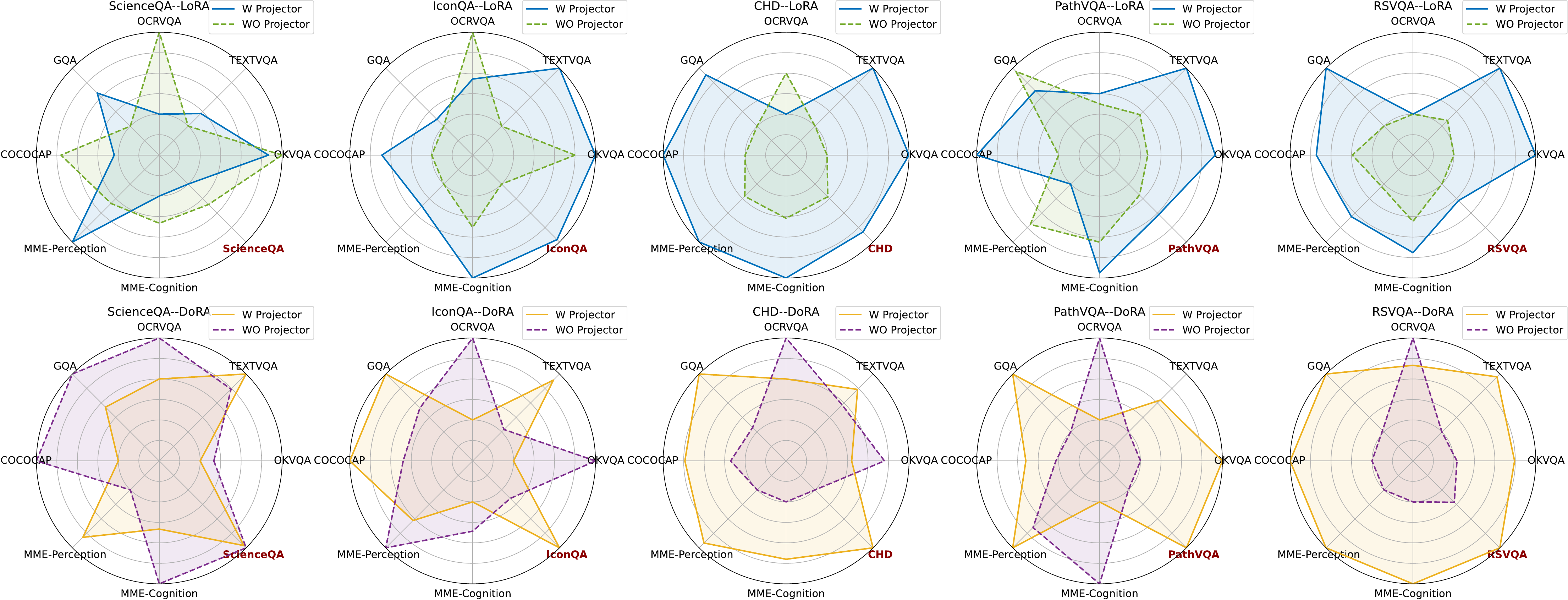}
\vspace{-10pt}
\caption{
{
\textbf{Performance Comparison on both Upstream and {\color{darkred}{Downstream}} tasks} with or without Vision Projector $\varphi$. Tuning projector benefits those distinct target distribution, \eg, \pathvqa{}, and \rsvqa{}. Refer to \cref{sec:exp_com} for discussion.}
}  
\label{fig:radar_performance}
\vspace{-10pt}
\end{figure*}

\subsection{Experimental Comparison}
\label{sec:exp_com}
In this section, we comprehensively evaluate different tuning methods for \mllm{} by addressing the following key questions.

\begin{fullitemize}
\item \textbf{\hypertarget{Q1}{Q1: Task specialization.}} Which methods generally achieve satisfying downstream performance?

\item \textbf{\hypertarget{Q2}{Q2: Stabilization resilience}} Whether existing tuning methods could maintain general ability? 

\item \textbf{\hypertarget{Q3}{Q3: Specialization and Stabilization Trade-Off.}} Do existing tuning solutions face the downstream and upstream performance balance dilemma?  

\item \textbf{\hypertarget{Q3}{Q4: Visual Adapter Uniqueness.}} What the unique character for visual projector model in tuning process.
\end{fullitemize}
Generally, we conduct experiments on both \llavaov{} and \vila{} architectures across different downstream datasets, as presented in \cref{tab:compare_sota_llava,tab:compare_sota_vila,fig:radar_performance}. 

\noindent To address \myhyperlink{Q1}{\textbf{Q1}}, we evaluate both downstream performance $\mathcal{A}_\mathcal{T}$ and specialization improvement $\mathcal{E}$. The results indicate that fully tuning all \llmabbrv{} blocks, referred to as \fullselft{} (\fullselftabbrv), achieves the highest task-specialization capability. However, for open-response tasks, both \lora{} and \fullselftabbrv{} suffer from severe overfitting, leading to performance degradation. By contrast, selective tuning of only the Top and Last layers shows enhanced specialization capability while effectively mitigating overfitting. We further observe that \lastselft{} (\lastselftabbrv{}) generally yields limited downstream performance, while \lora{} and \dora{} exhibit substantially poorer results on the CHD private dataset (see \cref{tab:compare_sota_vila}). Overall, \fullselft{} achieves consistent specialization performance, whereas Top-layer tuning provides comparable efficacy with reduced computational overhead.

\noindent In response to \myhyperlink{Q2}{\textbf{Q2}}, we assess the source distribution performance as $\mathcal{A}_{\mathcal{S}_i}$ and quantify stabilization forgetting via $\mathcal{F}$. The low-rank adaptation in \lora{} minimizes parametric conflict with original structures, thereby better preserving upstream generalization compared to \fullselft{}. Moreover, \lastselft{} (\lastselftabbrv{}) exhibits greater generalization degradation than \topselft{} (\topselftabbrv{}). These findings suggest that \lora{} and Top-layer tuning offer superior stabilization resilience through enhanced generalization retention.

\noindent Regarding \myhyperlink{Q3}{\textbf{Q3}}, \cref{tab:compare_sota_llava,tab:compare_sota_vila} reveal the inherent complexity in balancing specialization and stabilization, requiring simultaneous preservation of upstream capabilities and enhancement of downstream performance. Empirical results demonstrate that \lora{} and \topselftabbrv{} achieve relatively satisfactory equilibrium performance. In contrast, \fullselft{}, which updates all parameters within \llmabbrv{} blocks, leads to catastrophic forgetting of general knowledge, ultimately compromising model stability.

\noindent Regarding \myhyperlink{Q4}{\textbf{Q4}}, \cref{fig:radar_performance} visualizes the performance across both upstream and downstream datasets. We examine the impact of the visual projector through inclusion/exclusion of trainable connector modules ( $\varphi$) from the set of trainable parameters. Our findings suggest that tuning the projector adaptation generally enhances specialization for tasks with significant distribution shifts. However, for target distributions with similar characteristics to upstream, such as \scienceqa{}, freezing the connector often yields better results.

\subsection{Tuning Principe}
\label{sec:tuning_principe}
Driven by the experimental analysis in \cref{sec:exp_com}, we derive the following tuning principles to guide the selection of appropriate tuning methods and provide a deeper understanding of the underlying rationale. These insights aim to accelerate and enhance the application of \mllm{} across various downstream tasks:

\definecolor{keywordcolor}{RGB}{178,34,34}
\definecolor{c2framework}{RGB}{250, 249, 222} 
\begin{mdframed}[backgroundcolor=c2framework!40, linewidth=2pt, linecolor=c2framework!250, roundcorner=20pt]
\textbf{\textcolor{keywordcolor}{Tuning Principle}}: 
 
\noindent \textbf{\hypertarget{P1}{P1: Full Tuning Faces an Generalization and Specialization Balance Dilemma}}. Optimizing all parameters improves task-specific performance but risks overfitting, highlighting a critical trade-off between generalization and specialization in model optimization

\noindent \textbf{\hypertarget{P2}{P2: \lora{} Mitigates Catastrophic Forgetting but Limits Distribution Adaptation}}: Low-Rank Adaptation (\lora{}) preserves pretrained knowledge better than full tuning  but under-performs on distinct data distributions due to constrained parameter scalability.

\noindent \textbf{\hypertarget{P3}{P3: Top \llmabbrv{} Layers Encode Vision-Text Interaction}}: Tuning the top \llmabbrv{} layers (closest to the input) strengthens cross-modal alignment  but disrupts the model original visual-textual interaction dynamics.

\noindent \textbf{\hypertarget{P4}{P4: Final \llmabbrv{} Layers Focus on Output Style Imitation}}: Tuning the final \llmabbrv{} layers (closest to the output) enforces superficial style replication  while neglecting inherent distribution alignment learning.

\noindent \textbf{\hypertarget{P5}{P5: Vision Projector Adapts for Visual Shift Transfer}}: Adapting the vision projector addresses visual domain shifts but degrades the original encoder task-agnostic spatial-semantic representation.

\vspace{-5pt}
\end{mdframed}

To be precise, we provide a detailed analysis of the aforementioned tuning principles. For \myhyperlink{P1}{\textbf{P1}}, \fullselft{} (\fullselftabbrv{}) in \mllmabbrv{} typically involves training all parameters within the \llmabbrv{} module. As a result, \fullselftabbrv{} often achieves strong performance on downstream tasks. However, enabling full parameter updates introduces extensive flexibility, which can lead to overfitting on the target distribution. This overfitting results in a significant decline in generalization ability ($\mathcal{F}$), as demonstrated in \cref{tab:compare_sota_llava,tab:compare_sota_vila}. Therefore, \fullftabbr{} presents a double-edged sword, requiring a trade-off between specialization and generalization. Regarding \myhyperlink{P2}{\textbf{P2}}, \lora{} typically exploits the inherent low-rank matrix principle to adapt to the target distribution. However, it empirically underperforms compared to \fullftabbr{} across various tasks ($\mathcal{A}_{\mathcal{T}}$). Nonetheless, in terms of generalization ability, \lora{} mitigates conflicts with the original parameter space, thereby reducing source-domain forgetting ($\mathcal{F}$), as confirmed in related studies~\cite{LoraLearnForget_TMLR24,LoRAvsFull_arXiv24}. When the target task exhibits high visual distribution discrepancies, such as \rsvqa{} and \pathvqa{}, tuning the Top layers results in a more pronounced decline in upstream VQA performance compared to tuning the Last layers. This leads us to \myhyperlink{P3}{\textbf{P3}}, suggesting that the Top layers primarily facilitate vision-text interactions, and under significant distribution shifts, modifying them disrupts the original interaction patterns. Furthermore, we note that \cococap{} belongs to the open-response task category, requiring the model to generate descriptive captions. The effectiveness of \cococap{} tuning reflects its ability to adapt output style. However, tuning the Last layers in \llmabbrv{} leads to more severe performance degradation, a trend that is further amplified in \flickrthirtyk{}, another image captioning task. Consequently, we derive \myhyperlink{P4}{\textbf{P4}}, positing that tuning the Last \llmabbrv{} blocks primarily refines output text style without truly comprehending the multi-modal input or the underlying question, aligning with findings from~\cite{LimitofInstructionTuning_ICML24,REMEDY_ICLR25}. Lastly, we investigate the role of the visual projector. \cref{fig:radar_performance} presents the results for the \lora{} family, both with and without the visual projector module. The findings indicate that tuning the visual projector primarily benefits tasks exhibiting significant distributional divergence from the pre-training data, \eg, \pathvqa{}, \rsvqa{}. Based on these observations, we hypothesize that the visual projector plays a crucial role in transferring visual distribution shifts, thereby enhancing the model’s adaptation to the target distribution.

\section{Outlook}
\label{sec:outlook}

\subsection{Future Direction}
\label{sec:future_direction}

\subsubsection{Federated \mllmabbrv{} Tuning}
With respect to tuning \mllm{} (\mllmabbrv{}), it typically requires large-scale, high-quality datasets, which are both time-consuming and labor-intensive to obtain. Federated Learning, however, offers a feasible solution by enabling collaborative learning across decentralized parties without the need to exchange private data \cite{FedAvg_AISTATS17, FLOptimization_arXiv16, FLwithNonIID_arXiv18, FederatedMLConApp_TIST19, FLonNONIDD_NC21, FLSurveyandBenchmarkforGenRobFair_PAMI24}. Despite its advantages, transferring the full \mllmabbrv{} model incurs substantial communication costs due to its large scale. Therefore, investigating parameter-efficient Federated \mllmabbrv{} tuning methods is a crucial research direction {\cite{FedLLM_NeurIPS24,FLoRA_NeurIPS24,Openfedllm_SIGKKD24}}. These methods can effectively establish a privacy-preserving multi-party collaboration framework, enabling collaborative performance improvements on the target distribution. Additionally, the inherent data heterogeneity across distributed datasets leads to challenges such as divergent optimization directions, which slows down the convergence speed \cite{FCCL_CVPR22, FPL_CVPR23, HeteroFL_ICLR21, FedFA_ICLR23, FedFAPLUS_PAMI24, FedAS_CVPR24}. Consequently, designing effective federated global signals becomes a crucial problem to mitigate multi-client drift.

\subsubsection{Large and Small \mllmabbrv{} Collaboration}
\mllm{} (\mllmabbrv{}) have demonstrated exceptional performance in cross-modal perception, understanding, and interaction tasks. In recent years, edge devices such as mobile phones, smart wearable, and IoT sensors have become increasingly prevalent, enabling easy access to multi-modal sensory data. This trend has spurred significant interest in migrating \mllmabbrv{}-powered applications from centralized cloud infrastructures to the network edge {\cite{EDgeAIfor6G_JSAC21,LLMforAutoEdgeAI_CommMag24}}. However, the computational heterogeneity of edge devices—characterized by varying resource capabilities—necessitates the deployment of \mllmabbrv{}s at multiple scales. Consequently, there is a growing demand to establish collaborative tuning frameworks between large and small \mllmabbrv{} variants {\cite{AIFlow_Network25}}. While a straightforward solution involves aligning output distributions across models, this approach offers limited knowledge transfer potential {\cite{DynamicNNSurvey_PAMI21,DynamicTransformer_NeurIPS21,GFNet_PAMI22,EfficienttrainPlus_PAMI24,SelectiveFederatedDistillation_NC24}}. Therefore, developing an efficient collaborative paradigm for large and small \mllmabbrv{}s could significantly enhance their adaptability to diverse computing environments.

\subsection{Conclusion}
\label{sec:conclusion}
To our knowledge, this is the first work to comprehensively review recent advancements in \mllm{} tuning from the perspectives of Task-Expert Specialization and Open-World Stabilization.
We provide the  background knowledge and categorize over 100 \mllmabbrv{} tuning methods based on various criteria, including task settings, learning strategies, and technical contributions, encompassing \selft{}, \addft{}, and \repft{}.
Additionally, we present benchmarking results across six downstream datasets, including  medical analysis, remote sensing, and scientific knowledge.
Our discussion offers insights into key findings, open challenges, and promising future research directions in this field.
In conclusion, while \mllmabbrv{} has made remarkable progress due to rapid research advancements, achieving the specialization and generalization balance remains a significant challenge.

\ifCLASSOPTIONcaptionsoff
  \newpage
\fi



%
\vspace{-4pt}
{\small
\bibliographystyle{IEEEtran}

\bibliography{egbib}
}

\end{document}